\PassOptionsToPackage{dvipsnames}{xcolor} 

\documentclass[letterpaper]{article} %
\usepackage[preprint]{aaai2027}  %
\usepackage[hyphens]{url}  %
\usepackage{graphicx} %
\usepackage{natbib}  %
\usepackage{caption} %
\usepackage{soul}
\usepackage[symbol]{footmisc}
\newcommand{\our}{TGSR\xspace}
\newcommand{\cmark}{\textcolor{green!70!black}{\ding{51}}}
\newcommand{\xmark}{\textcolor{red}{\ding{55}}}

\usepackage{eccvabbrv}

\usepackage{graphicx}
\usepackage{siunitx, colortbl}
\usepackage{booktabs}
\usepackage{adjustbox}
\usepackage{multirow} 
\usepackage{arydshln}
\usepackage{pifont}

\usepackage{algorithm}
\usepackage{algorithmic}

\usepackage{newfloat}
\usepackage{listings}
\DeclareCaptionStyle{ruled}{labelfont=normalfont,labelsep=colon,strut=off} %
\floatstyle{ruled}
\newfloat{listing}{tb}{lst}{}
\floatname{listing}{Listing}

\usepackage{amsmath}
\usepackage[capitalize]{cleveref}
\crefname{section}{Sec.}{Secs.}
\Crefname{section}{Section}{Sections}
\crefname{table}{Tab.}{Tabs.}
\Crefname{table}{Table}{Tables}

\usepackage[accsupp]{axessibility}  %

\title{TGSR: Text-Grounded Diffusion for Faithful Image Super-Resolution}
\author{Zhiming Hu\textsuperscript{\rm 1},
Ran Zhang\textsuperscript{\rm 1*},
Angela Ye\textsuperscript{\rm 1*},
Raghav Goyal\textsuperscript{\rm 2},
Faraz Ali\textsuperscript{\rm 1},\\
Tristan Ty Aumentado-Armstrong\textsuperscript{\rm 1},
Samrudhdhi B. Rangrej\textsuperscript{\rm 1},\\
Konstantinos G. Derpanis\textsuperscript{\rm 3,4,5},
Iqbal Mohomed\textsuperscript{\rm 1$\dagger$},
Alex Levinshtein\textsuperscript{\rm 1$\dagger$}
}

\affiliations{
\textsuperscript{\rm 1} AIC-Toronto, Samsung Electronics
\textsuperscript{\rm 2} AMD
\textsuperscript{\rm 3} University of Toronto \\
\textsuperscript{\rm 4} Vector Institute 
\textsuperscript{\rm 5} York University 
\textsuperscript{*} Equal contribution
\textsuperscript{$\dagger$} Equal advising
}

\begin{document}

\maketitle

\begin{abstract}
Diffusion models have achieved remarkable success in image super-resolution (SR) by leveraging strong generative priors to enhance perceptual quality. However, for textual content, these priors often produce \textit{text hallucinations}: non-existent or distorted text, which alter scene semantics and compromise model reliability. While recent methods focus on enhancing text generation, they do so indiscriminately, leaving hallucinations fundamentally unresolved. We argue the core problem is a lack of selective control. Therefore, we reformulate text SR as a selective generation problem: text is resolved only when reliable generation is expected, while other scene content is fully super-resolved. We introduce TGSR (Text-Grounded Super-Resolution), a test-time framework that progressively identifies and suppresses likely hallucinated text regions during generation. To enable this, we train dedicated detectors on a large-scale dataset curated from diverse diffusion-based SR outputs. TGSR consistently improves text fidelity across multiple backbones, increasing recognition precision by up to 17.68 points without degrading perceptual quality, underscoring the value of selective control in generative SR.

\end{abstract}

\section{Introduction} \label{sec:intro}
\begin{figure*}
  \centering
  \includegraphics[width=1.0\textwidth]{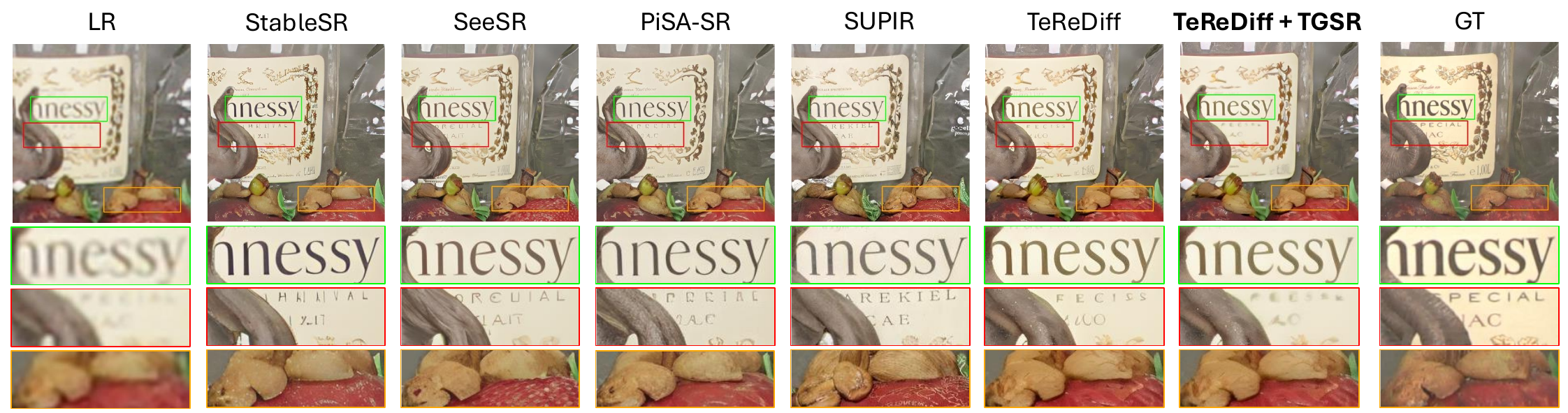}
  \caption{\textbf{Comparison of state-of-the-art approaches with \our (ours) on text-aware image super-resolution (SR)}. Existing SR models often produce distorted text when the corresponding regions in the low-resolution (LR) image are heavily degraded, as shown in the red bounding box. Our approach extends existing SR models (\eg, TeReDiff~\cite{min2025tair}) to suppress \textcolor{red}{hallucinated text} artifacts at inference time without sacrificing the visual quality of \textcolor{green}{legible text} and \textcolor{orange}{background} regions.} 
  \label{fig:teaser}
\end{figure*}

Diffusion models have achieved remarkable success in image super-resolution (SR) by leveraging strong generative priors to enhance perceptual quality~\cite{wu2024seesr, yang2024pasd, sun2024pisasr, wang2024exploiting,yu2024scaling}. However, this same generative strength introduces a critical limitation for textual content: diffusion-based SR models often generate text that does not exist in the ground truth image (see \cref{fig:teaser}), a phenomenon referred to as text hallucination~\cite{ren2025hallucination, min2025tair}. Such errors can arise from two sources: when the generative prior overrides insufficient evidence in the low-resolution (LR) input, or when the content falls outside the prior distribution (e.g., unseen fonts, languages, scripts, rare signage layouts, or unusual typographic patterns), leading to semantically incorrect content and altered scene meaning~\cite{aithal2025understanding, kim2025unified}. In both cases, hallucination emerges when generation is applied indiscriminately in regions where reliable character recovery cannot be confidently inferred from the available visual evidence and model predictions.

Mitigating text hallucination is therefore essential. Incorrect or fabricated text can fundamentally alter scene semantics, undermining reliability in applications where textual fidelity is crucial. Unlike generic perceptual artifacts, textual errors are discrete and semantically meaningful, making them immediately noticeable to humans and potentially misleading for downstream automated systems. Reducing hallucinated text is thus necessary for the trustworthy deployment of diffusion-based SR systems.

The most closely related work is text-aware SR~\cite{min2025tair, kim2025unified}. They detect character-level text in intermediate diffusion outputs and use the recognized content as conditioning signals during generation. While promising, they primarily focus on enhancing text generation itself and do not explicitly model the reliability of textual recovery, which can magnify hallucination artifacts in unreliable regions. Moreover, they typically finetune the backbone, making integration into arbitrary SR models difficult.

The core issue is the lack of explicit reliability modeling in generative refinement for textual recovery. Diffusion-based SR enhances all regions through generative refinement, implicitly assuming that stronger generation uniformly improves restoration. While this objective is effective for generic perceptual enhancement, it is fundamentally misaligned with the semantic nature of textual content. For text, the goal should not be to super-resolve every region indiscriminately, but to condition refinement on whether reliable character recovery can be confidently inferred. When reliability is anticipated, text can be confidently resolved; otherwise, further generative refinement risks introducing semantic errors. In such cases, preserving LR-consistent structure is preferable to introducing incorrect content. 

We therefore reformulate text-aware SR as a selective generation problem. Text is resolved only when reliable generation is expected, while the remaining scene content is fully super-resolved using diffusion priors. Guided by this principle, we introduce TGSR (Text-Grounded Super-Resolution), a test-time framework that progressively anticipates hallucinated text regions during SR generation and suppresses their emergence without retraining the underlying backbone. Concretely, at intermediate stages of the denoising process, we reuse the SR model's own diffusion features to detect regions likely to contain hallucinated text, adding only a lightweight detection head on top of the frozen backbone. The detections are then used for test-time adaptation, steering unreliable text regions toward a reference (a non-diffusion upsampled LR) image, while allowing all other regions to be super-resolved.

To support this selective control mechanism, we curate the first large-scale dataset for hallucinated text region detection in super-resolved images, termed SA-Text-HTD. By generating outputs from multiple diffusion-based SR backbones under diverse degradation settings, we obtain over 3.2M cropped text instances. We then leverage a multimodal large language model (MLLM) to automatically recognize the text in those cropped text instances, which are further compared against ground-truth transcriptions to produce hallucination labels at scale, enabling robust training and evaluation of hallucinated text detection. We commit to releasing the annotations and code to reproduce our dataset.

Our contributions are summarized as follows:
\begin{itemize}
\item  We reformulate text super-resolution as a selective generation problem to address hallucinated text.
\item  We define the novel task of hallucinated text region detection in SR outputs and introduce robust detection models for this purpose.
\item  We propose a test-time adaptation strategy that suppresses hallucinated text during inference without retraining the base SR model and without harming perceptual quality.
\item We curate a large-scale dataset for hallucinated text region detection -- using fully automated means -- to support training and evaluation.

\end{itemize}

Extensive experiments demonstrate that our method significantly improves text fidelity while preserving overall perceptual quality across multiple diffusion-based SR backbones. More broadly, our results highlight the importance of selective control in generative super-resolution.

\begin{figure*}[tb]
  \centering
  \includegraphics[width=\textwidth]{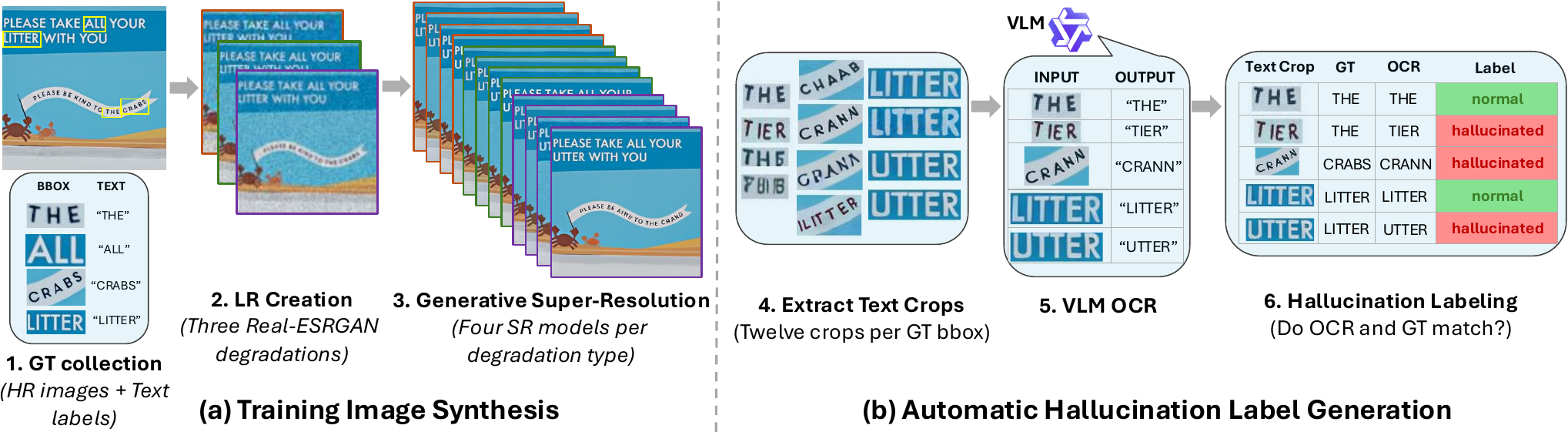}
  \caption{\textbf{Dataset curation pipeline.} From each ground-truth (GT) image, we generate three degradation levels using the Real-ESRGAN degradation adopted in prior works (\ie, TeReDiff, StableSR, PASD). The degraded images are super-resolved using four diffusion-based SR models: TeReDiff, PiSA-SR, PASD, and SUPIR. We then crop text regions based on the GT bounding boxes and recognize them using an MLLM-based OCR model. If the recognized text matches with GT transcription, the sample is labeled as \textbf{normal text}; otherwise, it is labeled as \textbf{hallucinated text}.}
  \label{fig:label_generation}
\end{figure*}

\section{Related Work}

\noindent
\textbf{Scene Text Image Super-Resolution (STISR).}
STISR aims to enhance LR text images to improve both visual quality and optical character recognition (OCR) accuracy. Existing methods typically focus on localized character reconstruction within cropped regions, often leveraging recognition guidance or generative priors~\cite{mou2020plugnet, chen2021scene, singh2024dcdm, luo2025restore, zhang2024diffusion, wang2020scene, ma2022text, ma2023text}. Unlike these localized methods, we focus on reliability-aware correction within full-image generative SR.

\noindent
\textbf{Diffusion-Based Image Super-Resolution.}
While diffusion-based SR has significantly improved realism through scaled capacity and efficient sampling~\cite{wang2024exploiting, yang2024pasd, wu2024seesr, lin2024diffbir, noroozi2024you, wu2024one, moser2024diffusion, sun2024pisasr, chen2025faithdiff, yu2024scaling}, most approaches apply generative refinement uniformly across the image. This lack of reliability modeling often leads to the hallucination of incorrect characters where underlying signals are ambiguous~\cite{min2025tair, kim2025unified}. We depart from this paradigm by incorporating hallucinated text region detection, providing an explicit reliability signal that allows for selective correction of evolving SR predictions.

\noindent
\textbf{Text-Aware Super-Resolution.}
Text-aware SR jointly trains OCR and diffusion components, using recognized content as conditioning~\cite{min2025tair, kim2025unified, hu2025text}. These methods improve recognition accuracy but require finetuning the diffusion backbone and restore all detected text indiscriminately, which can amplify hallucinations in unreliable regions. Instead, we reformulate text SR as a reliability-aware restoration problem. Rather than uniformly enhancing all regions, we selectively super-resolve text only where reliable character recovery is anticipated from intermediate SR outputs; regions predicted to be unreliable are preserved.

\section{Hallucinated Text Detection Dataset}

\subsection{Data Curation}
As no existing datasets target hallucinated text detection (HTD), we curate a new dataset, SA-Text-HTD, based on the SA-Text dataset~\cite{min2025tair}. We present an overview of our data curation pipeline in Fig.~\ref{fig:label_generation}. 

\noindent\textbf{Training Image Synthesis.}
We begin with the high-resolution (HR) images from SA-Text and apply Real-ESRGAN-based degradations~\cite{wang2021realesrgan} to generate low-resolution (LR) inputs. To ensure diverse distortion patterns, we adopt three degradation setups commonly used in prior work, namely, TeReDiff~\cite{min2025tair}, StableSR~\cite{wang2024exploiting}, PASD~\cite{yang2024pasd}. For each degraded LR image, we apply four diffusion-based SR models, namely TeReDiff~\cite{min2025tair}, PiSA-SR~\cite{sun2024pisasr}, PASD~\cite{yang2024pasd}, and SUPIR~\cite{yu2024scaling}, to obtain super-resolved outputs.

\subsubsection{Automatic Hallucination Label Generation.} \label{sec:autolabel}
Using the ground-truth (GT) bounding box annotations provided in SA-Text, we crop individual text instances from the SR images. Each cropped text region is then fed into Qwen2.5-VL-72B~\cite{bai2025qwen25vltechnicalreport} for recognition. The model is instructed to output either the recognized text or ``N/A'' if no text is identifiable. We compare the OCR output with the corresponding GT transcription. If the prediction matches the GT, the instance is labeled as \textit{normal}; otherwise, it is labeled as \textit{hallucinated}.

\noindent \textbf{OCR Model Choice.}
For OCR, we adopt Qwen2.5-VL-72B due to its strong recognition capability in challenging low-quality scenarios (Yang et al. 2025). To validate this choice, we conduct a comparative study against other OCR systems. Specifically, we apply PASD (Yang et al. 2024) to super-resolve LR images degraded using the PASD Real-ESRGAN setup, yielding 271,226 cropped text instances from the resulting SR outputs. Running PaddleOCRv5 (Cui et al. 2025) and Gemma4-31B~\cite{gemma4-31b} within the same data curation pipeline shows that their hallucination labels align with those produced by Qwen2.5-VL-72B on 94.32\% and 97.12\% of the samples, respectively. For the disputed crops between Qwen2.5-VL-72B and PaddleOCRv5, Qwen2.5-VL-72B correctly recognizes 85.3\% of samples, outperforming PaddleOCRv5’s 14.6\% accuracy. This highlights Qwen2.5-VL-72B’s robustness in handling degraded text, and we therefore adopt it as the default OCR model in our annotation pipeline.

\begin{table}[tb]
\centering
\fontsize{9}{10}\selectfont %
\setlength{\tabcolsep}{2.1pt}
\begin{tabular}{l|c|cc|c}
\toprule
\textbf{Dataset} & \textbf{Task} & \textbf{Text} & \textbf{Halluci.} & \textbf{\# of images} \\
& & \textbf{Label} & \textbf{Label} & \\
\hline
SA-Text & Text-Aware SR & \cmark & \xmark & 105,330 \\
CTW1500 & Text Spotting & \cmark & \xmark & 1,000 \\
TotalText & Text Spotting & \cmark & \xmark & 1,255 \\
\textbf{SA-Text-HTD} & Halluci. Text Det. & \cmark & \cmark & \textbf{12 $\times$ 105,330} \\
\bottomrule
\end{tabular}
\caption{\textbf{Comparison with other datasets.} Halluci. denotes \emph{hallucination}. Text-aware SR and text spotting datasets~\cite{yuliang2017detecting,ch2017total} provide text labels but lack hallucination labels. Our dataset includes both text and fine-grained hallucination labels. Moreover, we enlarge the SA-Text~\cite{min2025tair} dataset by a factor of $4 \mathrm{~(SR~models)}\times 3 \mathrm{~(degradation~levels)}$.
}
\label{tab:compare_dataset}
\end{table}

\subsection{Dataset Analysis}

\noindent
\textbf{Comparisons with Existing Dataset.}
Existing datasets~\cite{Agustsson_2017_CVPR_Workshops, cai2019toward, wei2020component} designed for general SR tasks do not contain text or hallucination annotations.
As in~\cref{tab:compare_dataset}, text-aware SR~\cite{min2025tair} and text spotting datasets~\cite{yuliang2017detecting,ch2017total} provide text-level annotations (\eg, bounding boxes and transcriptions), but none include explicit hallucination labels on SR images, which indicate whether the generated text is semantically correct. In contrast, our dataset provides both text annotations and fine-grained hallucination labels, enabling the study of hallucinated text detection in generative SR outputs. Moreover, our dataset is substantially larger in scale. Specifically, we expand the  SA-Text dataset by a factor of 12 (4 SR~models and 3 degradation~levels). More statistics are in the supplement.

\begin{figure*}[htb]
  \centering
  \includegraphics[trim={0 9.7cm 2.7cm 0.05cm},clip,width=\textwidth]{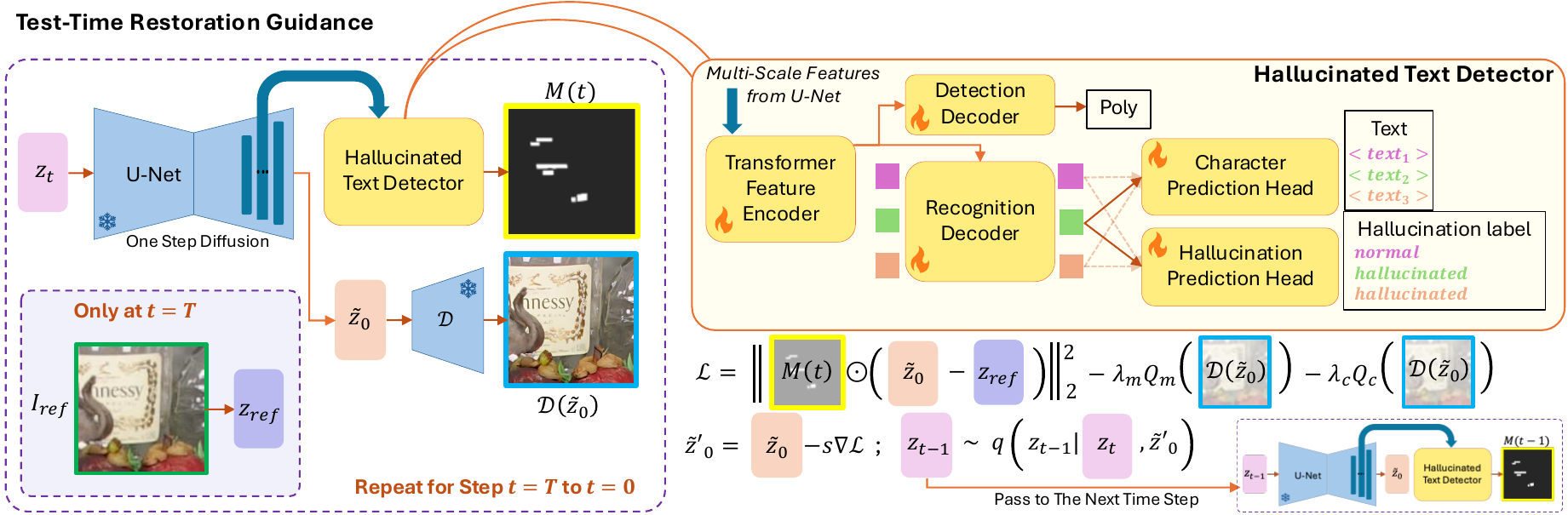}
  \caption{\textbf{Overall architecture of TGSR}. The framework contains two key components: a \textbf{hallucinated text detector} and \textbf{test-time restoration guidance}. 
  The detector reuses the multi-scale features from the SR model's own frozen denoising U-Net: only a lightweight text-spotting head is trained, which
  outputs text polygons with binary labels indicating whether each text instance is normal or hallucinated. During inference, at each diffusion step, we first predict the clean latent through denoising, 
  and apply the detector head to obtain hallucinated text masks. The masked regions are used to compute the loss, and a one-step gradient update is applied to the clean latent before predicting $z_{t-1}$. This process repeats for all $T$ diffusion steps.
  }
  \label{fig:detecor_and_tta}
\end{figure*}

\section{Methods}

The overall framework of TGSR is illustrated in \cref{fig:detecor_and_tta}. Our approach consists of two tightly coupled components: a \emph{hallucinated text detector} (\cref{sec:htd})  and a \emph{test-time restoration guidance} module (\cref{sec:mitigation}).

Given the multi-scale features computed by the SR model's denoising U-Net for a super-resolved image, $I_{\text{sr}}$, (or intermediate diffusion outputs),
the hallucinated text detector identifies hallucinated text instances and produces a spatial mask at time step $t$, denoted by $M(t)$, highlighting unreliable text regions. 

Conditioned on this mask, we introduce a test-time adaptation method that operates during the denoising process of diffusion-based SR models. The core idea is to selectively suppress hallucinated text by replacing it with high-fidelity evidence from a non-hallucinatory reference image, $I_{\text{ref}}$ (\eg, the SwinIR~\cite{liang2021swinir} output) in a harmonized manner, while leaving other regions unaffected. \our is model-agnostic and can be seamlessly integrated into diffusion-based SR pipelines without backbone finetuning.

\subsection{Hallucinated Text Detection}
\label{sec:htd}

Detecting hallucinated text requires both accurate text localization and a reliable assessment of whether each detected text instance is hallucinated. To this end, our hallucinated text detector adapts prior text detection work~\cite{Zhang_2022_CVPR}. First, following recent work~\cite{min2025tair}, we replace conventional backbone features (\eg, ResNet-50) with multi-scale diffusion features extracted from the four decoder blocks in the U-Net backbone of the diffusion model. 
Crucially, rather than equipping the detector with a diffusion backbone of its own, we extract these features from the SR model's \emph{own} frozen U-Net: detection reuses the features already computed during SR generation, so only a lightweight text-spotting head is trained for each backbone and no separate feature-extraction network is needed at inference.

The feature maps from each scale are first processed by separate convolutional layers to align their channel dimensions. Each map is then flattened into a sequence of tokens, and fed into a transformer-based feature encoder. The encoder predicts coarse bounding boxes and confidence scores under the supervision of a text classification loss, bounding box regression loss, and generalized IoU loss~\cite{rezatofighi2019generalized}. The encoded features are further used by a \emph{detection decoder} for polygon prediction and a \emph{recognition decoder} for character recognition. Unlike the encoder losses, these decoders are supervised using a text classification loss, polygon regression loss ($L_1$), and character recognition loss (cross-entropy), following previous work~\cite{Zhang_2022_CVPR, min2025tair}; detailed losses are listed in the supplement.

Unlike standard text spotting models that focus solely on localization and recognition, our task additionally requires distinguishing hallucinated text from normal text. We therefore augment the architecture with a dedicated \emph{hallucination prediction head}. It takes the character-level features from the recognition decoder as input, leveraging character-level representations to better identify text distortions. It outputs character-level logits, which are averaged into text instance-level hallucination predictions.

During training, we optimize the detector using the aforementioned encoder and decoder losses, and introduce an additional focal loss~\cite{lin2017focal} for the hallucinated text prediction head to address class imbalance. For the $j$-th query, the hallucinated text classification loss is defined as:
\begin{align}
\mathcal{L}_{\mathrm{halluci\_cls}}^{(j)} &= - \mathbf{1}_{\{j \notin \mathrm{Im}(\sigma)\}} \, (1 - \alpha) (\hat{b}^{(j)})^{\gamma} \log(1 - \hat{b}^{(j)}) \notag\\
& - \mathbf{1}_{\{j \in \mathrm{Im}(\sigma)\}} \, \alpha (1 - \hat{b}^{(j)})^{\gamma} \log(\hat{b}^{(j)})
\end{align}
where $\hat{b}^{(j)}$ is the predicted confidence score, $\gamma$ is the focusing parameter in the focal loss, $\alpha$ is a class balancing parameter, and $\text{Im}(\sigma)$ is the set of prediction indices matched to GT instances under optimal bipartite assignment, $\sigma$. 

\subsection{Mitigating Text Hallucinations}
\label{sec:mitigation}
Once hallucinated text regions are identified, the next challenge is how to effectively mitigate them. This problem poses three key difficulties: (i) defining a meaningful optimization direction that suppresses hallucinations, (ii) restricting modifications to hallucinated regions while preserving overall perceptual quality, and (iii) achieving these goals without finetuning powerful pretrained generative SR models.

We address these challenges via a test-time adaptation strategy applied during the diffusion denoising process, which can be integrated into existing diffusion-based SR models without additional training. At each denoising step, $t$, the diffusion model predicts a clean latent, $\tilde{z}_0$. 
The multi-scale U-Net features from the same denoising pass are fed to the hallucinated text detector to identify the hallucinated text regions.
The detector produces a binary mask, $M(t)$, indicating hallucinated text regions, where masked pixels are set to one and all others to zero.

To enforce reliable restoration, we leverage a reference image $I_{\text{ref}}$ (\ie, a non-generative SR reference) that captures the information in the LR image to define a region-aware reconstruction loss that encourages consistency between the predicted clean latent, $\tilde{z}_0$, %
and $I_{\text{ref}}$ within the hallucinated text regions only. Specifically, we compute a masked mean squared error (MSE) loss in latent space between $\tilde{z}_0$ and the encoded reference, $\mathcal{E}(I_{\text{ref}})$, where $\mathcal{E}$ is the VAE encoder,  
to steer the hallucinated text regions toward the reference content, which typically is reliable in content but lacking in perceptual quality. In addition, we incorporate two no-reference perceptual terms, MUSIQ~\cite{ke2021musiq} and CLIPIQA~\cite{wang2023exploring}, computed on the decoded image, $\mathcal{D}(\tilde{z}_0)$, where $\mathcal{D}$ is the VAE decoder,
to stabilize global perceptual quality and prevent degradation outside the masked regions. Formally, the loss is defined as

\begin{align}
    \mathcal{L}(\tilde{z}_0) ={} & \left\lVert M(t) \odot \big( \tilde{z}_0 - \mathcal{E}(I_{\mathrm{ref}}) \big) \right\rVert_2^2 \notag \\
    & - \lambda_m Q_m\big(\mathcal{D}(\tilde{z}_0)\big) - \lambda_c Q_c\big(\mathcal{D}(\tilde{z}_0)\big),
\end{align}
where $M(t)$ is downsampled to the latent resolution, $Q_m$ and $Q_c$ denote the MUSIQ and CLIPIQA models, and $\lambda_m$, $\lambda_c$ are the corresponding loss weights.

After computing the loss, we perform a single gradient descent step on $\tilde{z}_0$ to obtain an updated latent, $\tilde{z}_0'$, without updating model parameters:
\begin{equation}
\label{eq:gradient_descent}
    \tilde{z}'_0 = \tilde{z}_0 - s \nabla_{\tilde{z}_0} \mathcal{L}(\tilde{z}_0),
\end{equation}
where $s$ denotes the step size of the test-time adaptation update.
When the detector finds no hallucinated text at a given step, \ie, $M(t) = \mathbf{0}$, we keep the perceptual terms active and damp the gradient step by a factor $\rho < 1$.

The updated latent, $\tilde{z}_0’$, is then used to sample the next latent, $z_{t-1}$, from the posterior distribution, $q(z_{t-1} \mid z_t, \tilde{z}_0')$, following the standard reverse diffusion process. This procedure is repeated for all $T$ denoising steps until the final output, $z_0$, is obtained. By constraining optimization to hallucinated regions and operating solely at test time, our method effectively reduces text hallucinations while preserving the strong generative prior of the original diffusion model.

\section{Evaluation}
\subsection{Experimental Setting} 

{
\setlength{\tabcolsep}{0.01in}
\fontsize{9}{10}\selectfont
\aboverulesep = 0mm
\belowrulesep = 0mm 
\begin{table*}[t]
    \centering
    \sisetup{
        detect-weight, 
        mode=text,
        input-symbols={}, 
        table-format=2.3, 
        text-series-to-math = true,
        propagate-math-font = true,
    }
    \begin{tabular}{@{}lS[table-format=2.2]S[table-format=1.3]S[table-format=1.3]S[table-format=2.2]S[table-format=1.3]S[table-format=1.3]S[table-format=2.2]S[table-format=2.2]S[table-format=2.2]@{}}
    \toprule
      & {PSNR $\uparrow$} & {SSIM $\uparrow$} & {LPIPS $\downarrow$} & {MUSIQ $\uparrow$} & {CLIPIQA $\uparrow$} & {QAlign $\uparrow$} & {OCR-P $\uparrow$} & {OCR-R $\uparrow$ }& {OCR-F1 $\uparrow$ } \\
    \midrule
      Bicubic & 27.86 & 0.858 & 0.362 & 24.13 & 0.283 & 1.621 & 68.84 & 22.13 & 33.50 \\
      Swin2SR \cite{conde2022swin2sr} & 27.64 & 0.879 & 0.248 & 48.43 & 0.367 & 2.509 & 63.87 & 30.75 & 41.51 \\
      Real-ESRGAN \cite{wang2021realesrgan} & 25.69 & 0.846 & 0.231 & 58.58 & 0.580 & 3.228 & 58.85 & 30.43 & 40.11 \\
      StableSR~\cite{wang2024exploiting} & 26.44 & 0.826 & 0.264 & 57.12 & 0.592 & 3.026 & 58.10 & 36.30 & 44.68 \\
      PASD~\cite{yang2024pasd} & 26.15 & 0.831 & 0.265 & 57.01 & 0.528 & 3.208 & 59.81 & 39.90 & 47.87 \\
      SeeSR~\cite{wu2024seesr} & 25.69 & 0.829 & 0.279 & 67.09 & 0.654 & 3.667 & 54.94 & 37.75 & 44.75 \\
      FaithDiff~(\citeauthor{chen2025faithdiff}) & 25.75 & 0.795 & 0.297 & 64.94 & 0.604 & 3.486 & 56.88 & 39.85 & 46.87 \\
      DiT4SR~\cite{duan2025dit4sr} & 24.36 & 0.782 & 0.296 & 63.88 & 0.625 & 3.397 & 53.99 & 43.35 & 48.09 \\
     \cdashline{1-10}\noalign{\vskip 2pt}
      PiSA-SR~\cite{sun2024pisasr} & 26.46 & 0.851 & \textbf{0.241} & 64.37 & 0.611 & \textbf{3.260} & 54.71 & \textbf{40.98} & 46.86 \\
      \textbf{PiSA-SR +TGSR (ours)} & \textbf{26.52} & \textbf{0.852} & 0.244 & \textbf{64.72} & \textbf{0.615} & 3.212 & \textbf{68.33} & 40.66 & \textbf{50.98} \\
     \cdashline{1-10}\noalign{\vskip 2pt}
      SUPIR~\cite{yu2024scaling} & 24.16 & 0.706 & 0.394 & 59.58 & 0.659 & \textbf{3.481} & 44.94 & \textbf{36.56} & 40.32 \\
      \textbf{SUPIR +TGSR (ours)} & \textbf{24.27} & \textbf{0.709} & \textbf{0.392} & \textbf{60.81} & \textbf{0.663} & 3.463 & \textbf{51.29} & 36.40 & \textbf{42.58} \\
     \cdashline{1-10}\noalign{\vskip 2pt}
      TeReDiff~\cite{min2025tair} & 24.98 & 0.809 & 0.286 & 62.37 & 0.643 & \textbf{3.551} & 53.24 & \textbf{42.43} & 47.23 \\
      TeReDiff +TGSR (guide all text) & \textbf{26.13} & \textbf{0.833} & \textbf{0.258} & 61.85 & 0.606 & 3.061 & 65.64 & 37.75 & 47.93 \\
    \textbf{TeReDiff +TGSR (ours)} & 25.05 & 0.811 & 0.286 & \textbf{67.24} & \textbf{0.697} & 3.534 & \textbf{70.92} & 42.16 & \textbf{52.89} \\
    \bottomrule
    \end{tabular}
    \caption{
    \textbf{SR results on Real-Text dataset~\cite{min2025tair}.} We evaluate on FR-IQA, NR-IQA, and end-to-end text recognition metrics. We compare base SR models and our adapted models with test-time adaptation, as denoted by \textbf{+TGSR (ours)}. Our method increases recognition precision (OCR-P) and OCR-F1 while maintaining the perceptual quality.}
    \label{tab:real-text-main}
\end{table*}
}

\noindent
\textbf{Evaluation dataset for test-time adaptation.} For test-time adaptation, we evaluate on the Real-Text dataset and the SA-Text-Test dataset, following the setup in TeReDiff~\cite{min2025tair}. The Real-Text dataset contains 847 real-world HR-LR pairs extracted from RealSR~\cite{cai2019toward} and DRealSR~\cite{wei2020component}, with text instances annotated by~\citet{min2025tair}. SA-Text-Test~\cite{min2025tair} contains 1,000 HR images and their corresponding LRs generated with three levels (levels 1-3) of Real-ESRGAN degradations, as used in DASR~\cite{jie2022DASR}.

\label{sec:test_set_creation}
\noindent
\textbf{Training and evaluation dataset for HTD.} The hallucinated text detector is trained on our SA-Text-HTD dataset, which includes SR outputs from four diffusion-based SR models on three levels of degradations. In focal loss, $\alpha$ and $\gamma$ are set to be 0.25 and 2.0, respectively. For evaluation, we construct Real-Text-HTD from Real-Text~\cite{min2025tair} following \cref{sec:autolabel}.
For testing, we include SR outputs from TeReDiff, and \textit{two SR models unseen during detector training}, SeeSR~\cite{wu2024seesr} and StableSR~\cite{wang2024exploiting}, to benchmark robustness on unseen SR models.

\noindent
\textbf{Evaluation metrics.} 
For evaluating test-time guidance, we adopt widely used full-reference image quality assessment (FR-IQA) metrics (\eg, PSNR, SSIM~\cite{wang2004image}, LPIPS~\cite{zhang2018unreasonable}) and no-reference image quality assessment (NR-IQA) metrics (\eg, MUSIQ~\cite{ke2021musiq}, CLIPIQA~\cite{wang2023exploring} and QAlign~\cite{wu2024qalign}). To measure the text quality, we adopt text recognition metrics produced by off-the-shelf text spotting models (\ie, TESTR~\cite{Zhang_2022_CVPR}). We report  end-to-end recognition precision (OCR-P), recognition recall (OCR-R) and the harmonic mean of recognition precision and recall (OCR-F1). For the hallucinated text detector, we use the detection precision and recall on the hallucinated text regions. 

\noindent
\textbf{Implementation and training details}. We evaluate \our on three state-of-the-art diffusion-based SR baselines: TeReDiff~\cite{min2025tair} (text-aware SR model), PiSA-SR~\cite{sun2024pisasr} (single-step SR model), and SUPIR~\cite{yu2024scaling} (multi-step SR model). For each backbone, we freeze the diffusion U-Net and train only the text-spotting head on its multi-scale features, using the AdamW~\cite{loshchilov2017decoupled} optimizer with a learning rate of $1 \times 10^{-4}$. We classify a text region as hallucinated only if the predicted score exceeds a threshold of 0.3. For test-time guidance, we set $\lambda_m = 10^{-3}$, $\lambda_c = 10^{-4}$, and the step size $s$ in \cref{eq:gradient_descent} to 1 with normalized gradients; on steps where no hallucinated text is detected, the step is damped by $\rho = 0.3$. More details are in the supplement.

\subsection{Main Results}

\begin{figure*}[tb]
  \centering
  \begin{minipage}{\linewidth}
    \text{\scriptsize ~~~~~~~~~~~~GT~~~~~~~~~~~~~~~~~~~~~~~~~GT~~~~~~~~~~~~~~~~~~~Bicubic~~~~~~~~~~~~~~StableSR~~~~~~~~~~~~~~~~SeeSR ~~~~~~~~~~~~~~~~~~PASD~~~~~~~~~~~~~~~~~~PiSA-SR~~~~~~~~~~~~~~~~SUPIR~~~~~~~~~~~~~~~~~TeReDiff~~~~~~~~TeReDiff + Ours}
    \includegraphics[width=0.5\linewidth]{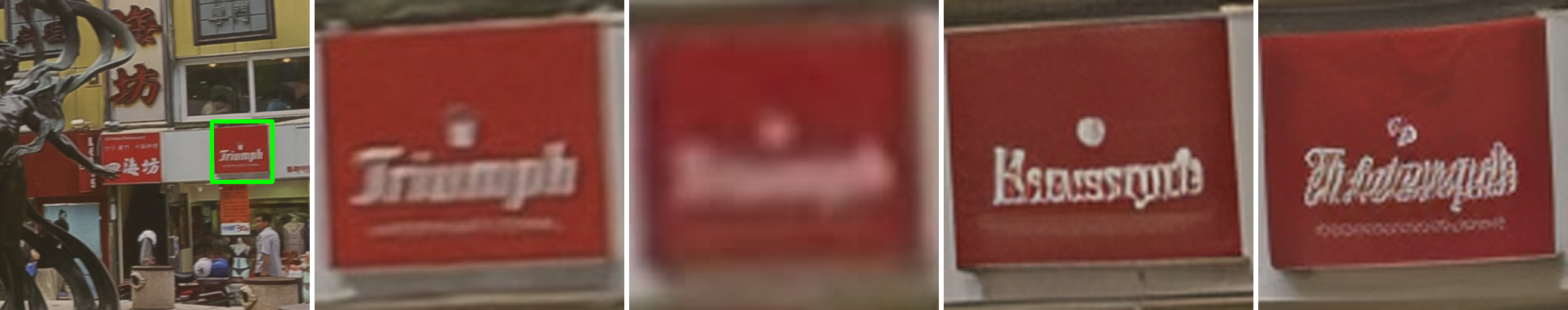}
    \hfill
    \includegraphics[width=0.5\linewidth]{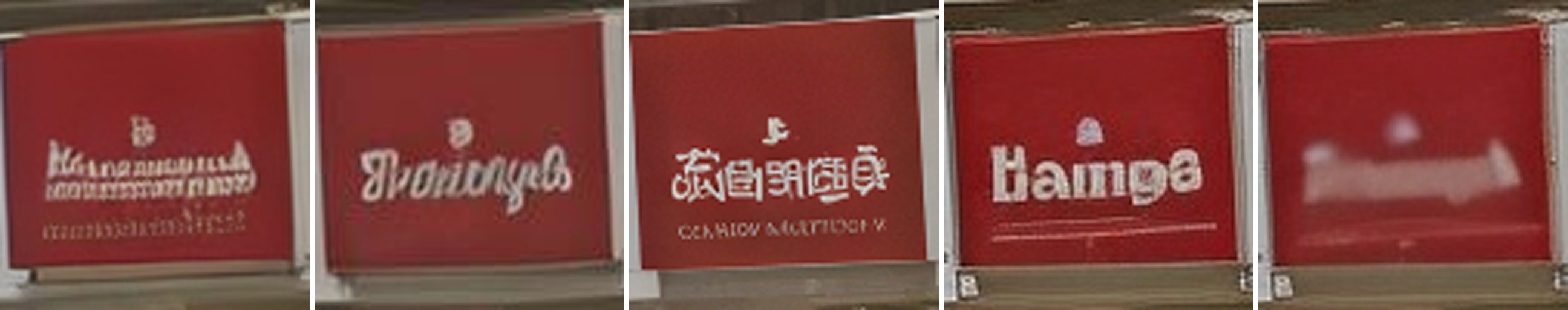}    
  \end{minipage}
  \vfill
  \begin{minipage}{\linewidth}
    \includegraphics[width=0.5\linewidth]{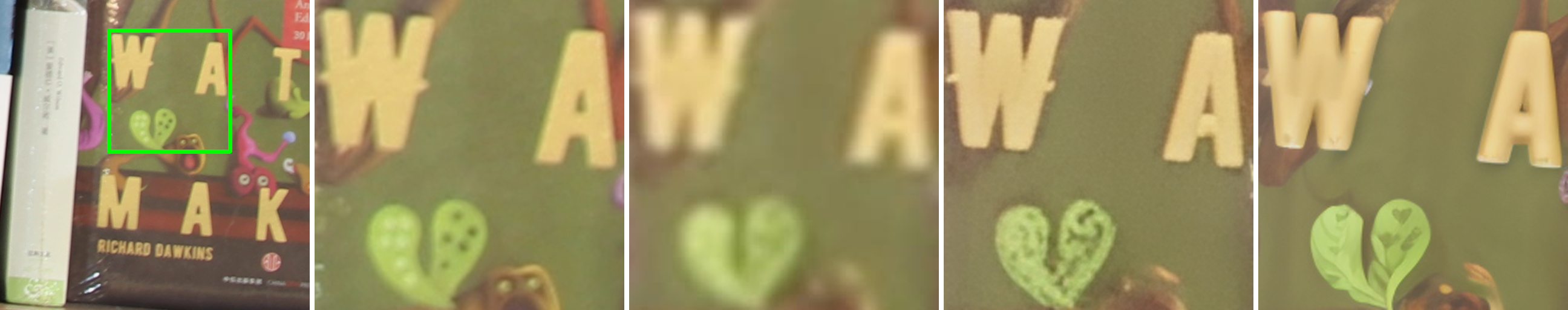}
    \hfill
    \includegraphics[width=0.5\linewidth]{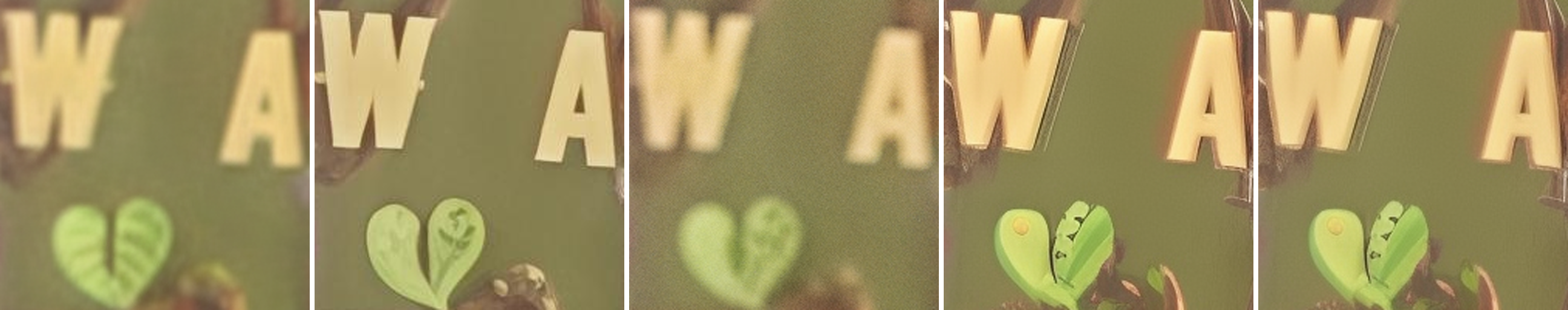}
  \end{minipage}
  \caption{\textbf{Qualitative results on the Real-Text dataset}. Our method suppresses text hallucinations (top) and preserves background details (bottom) compared to other diffusion-based SR models.}

  \label{fig:tgsr_qual}
\end{figure*}

\textbf{Test-Time Guidance}.
We report test-time adaptation results on the Real-Text dataset in \cref{tab:real-text-main}. TGSR consistently improves text recognition performance while preserving overall image quality. For example, incorporating TGSR into TeReDiff~\cite{min2025tair} improves the OCR-F1 score by 5.66 points (from 47.23 to 52.89), while improving NR-IQA metrics ($+4.87$ MUSIQ, $+0.054$ CLIPIQA), and achieving comparable QAlign and FR-IQA metrics.
Notably, OCR precision increases by 17.68 points with comparable recall.
Similar trends are observed when integrating TGSR with SUPIR~\cite{yu2024scaling} and PiSA-SR~\cite{sun2024pisasr}, where improvements are primarily driven by substantial gains in recognition precision. 
This improvement stems from our hallucinated text detector, which identifies unreliable text regions during denoising. 
For these regions, TGSR enforces consistency with the reference image instead of allowing the diffusion prior to generate distorted or hallucinated text-like patterns. 
Results on SA-Text-Test (supplement \cref{sec:result_tta_supp}) follow the same pattern: \our improves OCR-P and OCR-F1 for all three backbones at all three degradation levels while preserving image quality, with precision gains that grow with degradation severity (up to 20.7 points at level 3).

\noindent
\textbf{Is the Detector Necessary?} To separate the contribution of hallucinated text detection from that of reference substitution, we apply identical guidance to all spotter-detected text regions, without hallucination classification (\cref{tab:real-text-main}, guide all text). This raises OCR-P to 65.6 but costs 4.68 points of recall (42.43→37.75), leaving OCR-F1 at 47.93 — indistinguishable from the base model. \our instead reaches 70.92 precision at a recall cost of 0.27, dominating indiscriminate suppression on both axes. \cref{tab:htd_main} anticipates this: treating all detected text as hallucinated yields only 28.2 precision, so suppressing all of them inevitably degrades legible regions.

\begin{table}[t]
\centering
\fontsize{9}{10}\selectfont %
\setlength{\tabcolsep}{2.1pt}
\begin{tabular}{l ccc ccc ccc}
\toprule
& \multicolumn{3}{c}{TeReDiff} & \multicolumn{3}{c}{SeeSR$^{\dagger}$} & \multicolumn{3}{c}{StableSR$^{\dagger}$} \\
\cmidrule(lr){2-4} \cmidrule(lr){5-7} \cmidrule(lr){8-10}
Detector & P$\uparrow$ & R$\uparrow$ & F1$\uparrow$ & P$\uparrow$ & R$\uparrow$ & F1$\uparrow$ & P$\uparrow$ & R$\uparrow$ & F1$\uparrow$ \\
\midrule
TESTR & 28.2 & 47.7 & 35.4 & 24.2 & 34.4 & 28.4 & 23.5 & 30.4 & 26.5 \\
TeReDiff & 25.2 & \textbf{55.9} & 34.7 & 25.2 & \textbf{47.4} & 33.0 & 26.0 & \textbf{44.9} & 32.9 \\
\midrule
Ours (RN50) & 53.0 & 32.6 & 40.4 & 57.2 & 28.7 & \textbf{38.2} & 60.4 & 26.9 & \textbf{37.2} \\
Ours (DF)  & \textbf{60.1} & 32.6 & \textbf{42.3} & \textbf{62.4} & 24.7 & 35.4 & \textbf{65.3} & 25.5 & 36.7 \\
\bottomrule
\end{tabular}
\caption{\textbf{Hallucinated text detection on Real-Text-HTD.}
Precision (P) and recall (R) on hallucinated regions at the deployed
operating point ($\tau{=}0.3$); $\dagger$ marks SR backbones unseen
during detector training. In the context of test-time adaptation, false positives may erroneously blur sharp text by pulling it toward the reference image, whereas false negatives simply result in a no-op for that region. Consequently, we prioritize precision in the hallucinated text detector to avoid degrading existing high-quality textual content.}
\label{tab:htd_main}
\end{table}

\noindent\textbf{Hallucinated Text Detection.} Tab.~\ref{tab:htd_main} evaluates our detector on Real-Text-HTD. The detector is trained on the full SA-Text-HTD dataset, \ie, SR outputs from all four backbones under all three degradation settings, while Real-Text-HTD includes SR outputs from two backbones unseen during training. Ours (DF) utilizes the multi-scale diffusion features from TeReDiff and Ours (RN50) uses a ResNet-50 feature extractor.
Hallucinated text detection is a new and intrinsically hard task:
hallucinated regions are, by construction, those that most resemble
plausible text. To establish baselines, we adapt existing detectors (TESTR and TeReDiff) to treat all detected text instances as hallucinations. In this setup, our detector more than doubles the precision of both the state-of-the-art text spotter (TESTR) and the pretrained TeReDiff detector (60.1–65.3 vs. 24.2–28.2), resulting in a superior F1 score. This advantage holds on SR backbones unseen during detector
training, indicating that the detector generalizes across
diffusion SR outputs. Moreover, DF trades recall for precision relative to RN50.
 We deliberately
operate at a high-precision, moderate-recall point: a false positive
suppresses legible text and directly harms readability, whereas a
missed detection merely leaves the base model's output in place, so
precision governs the safety of test-time intervention. Crucially,
the detector need not be perfect to be useful, as \cref{tab:real-text-main} shows, this operating point already yields 6--18 points of OCR-P downstream while preserving
perceptual quality. More results on HTD are provided in the supplement.

\begin{table}[t]
\centering
\small
\setlength{\tabcolsep}{3.0pt}
\begin{tabular}{l cc cc cc}
\toprule
& \multicolumn{2}{c}{OCR-P$\uparrow$} & \multicolumn{2}{c}{OCR-R$\uparrow$} & \multicolumn{2}{c}{OCR-F1$\uparrow$} \\
\cmidrule(lr){2-3} \cmidrule(lr){4-5} \cmidrule(lr){6-7}
Text size & Base & +TGSR & Base & +TGSR & Base & +TGSR \\
\midrule
Small (p25--p50)      & 40.2 & \textbf{64.9} & 46.0 & 44.9 & 42.9 & \textbf{53.1} \\
Ext.\ small ($<$p25)  & 15.0 & \textbf{36.5} & 12.3 & 10.8 & 13.5 & \textbf{16.6} \\
\bottomrule
\end{tabular}
\caption{\textbf{Recognition by text size on Real-Text} (via
GT polygon area percentile), for TeReDiff (Base) and with TGSR.
Gains are largest exactly where hallucination is most frequent:
on extremely small text, precision more than doubles.}
\label{tab:textsize}
\end{table}

\noindent\textbf{Performance on the Hardest Text.}
Hallucinations concentrate in small text, where the LR evidence is
weakest. Tab.~\ref{tab:textsize} breaks down end-to-end recognition
by text size. TGSR's gains grow as conditions worsen: precision
improves by +24.7 points on small text and more than doubles
(15.0$\rightarrow$36.5) on extremely small text, at a modest recall
cost. Selective control is thus most valuable precisely where
indiscriminate generation is least reliable.%

\noindent
\textbf{Qualitative Study}. As shown in \cref{fig:tgsr_qual}, \our effectively suppresses hallucinations by identifying unreliable text regions and enforcing consistency with the reference image. 

\begin{table}[tb]
\centering
\fontsize{9}{10}\selectfont
\setlength{\tabcolsep}{2.0pt}
\begin{tabular}{l cc ccc}
\toprule
& \multicolumn{2}{c}{NR-IQA} & \multicolumn{3}{c}{OCR} \\
\cmidrule(lr){2-3} \cmidrule(lr){4-6}
Configuration & MUSIQ$\uparrow$ & CLIPIQA$\uparrow$ & P$\uparrow$ & R$\uparrow$ & F1$\uparrow$ \\
\midrule
TeReDiff          & 62.37 & 0.643 & 53.24 & 42.43 & 47.23 \\
\textbf{Ours}     & 67.24 & 0.697 & 70.92 & 42.16 & 52.89 \\
\midrule
\multicolumn{6}{l}{\emph{Guidance space and objective}} \\
RGB, MSE                  & 60.93 & 0.627 & 68.60 & 42.11 & 52.19 \\
Latent, MSE               & 61.39 & 0.631 & 70.93 & 41.79 & 52.59 \\
Latent, $+$\,MUSIQ        & 67.62 & 0.646 & 70.91 & 42.00 & 52.76 \\
Latent, $+$\,CLIPIQA      & 61.69 & 0.724 & 69.66 & 41.30 & 51.86 \\
\bottomrule
\end{tabular}
\caption{
\textbf{Ablation studies on Real-Text} using TeReDiff as the base model. Our default configuration employs a masked latent MSE objective, incorporates MUSIQ ($10^{-3}$) and CLIPIQA ($10^{-4}$) on decoded images, and applies a dampening factor of $\rho{=}0.3$ for empty masks. 
}
\label{tab:ablation_v3}
\end{table}

\subsection{Ablation Studies}
We conduct ablation studies on the Real-Text dataset (\cref{tab:ablation_v3}). Additional ablations are in the supplement. 

\noindent\textbf{Guidance space.}
We first compare applying the masked MSE guidance in RGB space versus latent space. Latent-space guidance performs better across the board, improving OCR-P by 2.3 points and OCR-F1 by 0.4 points at comparable NR-IQA scores, so we adopt it as our default.

\noindent\textbf{Guidance objective.}
On top of the latent MSE term, combining both MUSIQ and CLIPIQA, our default, improves both NR-IQA metrics over the MSE-only variant ($+5.9$ MUSIQ, $+0.066$ CLIPIQA) while achieving the best OCR-F1 (52.89), giving the best recognition--quality trade-off. 

\section{Conclusion}

In this work, we addressed text hallucination in diffusion-based image super-resolution.  The key idea is reformulating text SR as a selective generation problem.  We introduced TGSR, a test-time framework that identifies likely hallucinated text regions during generation and suppresses their emergence while allowing other scene elements to be fully super-resolved.  Experiments showed that TGSR improved text fidelity across multiple diffusion-based SR backbones without degrading perceptual quality.  Beyond text, our reliability-aware approach may extend to other important identity-sensitive visual elements, such as faces, license plates, barcodes, and QR codes.

\bibliography{main}

@String { AAAI         = {{Proceedings of the National Conference on Artificial Intelligence ({AAAI})}} }

@String { CVPR         = {{Proceedings of the {IEEE} Conference on Computer Vision and Pattern Recognition ({CVPR})}} }

@String { CVPRW        = {{Proceedings of the {IEEE} Conference on Computer Vision and Pattern Recognition Workshops ({CVPRW})}} }

@String { ECCV         = {{Proceedings of the European Conference on Computer Vision ({ECCV})}} }

@String { ICCV         = {{Proceedings of the International Conference on Computer Vision ({ICCV})}} }

@String { ICCVW         = {{Proceedings of the International Conference on Computer Vision Workshops ({ICCVW})}} }

@String { ICLR         = {{Proceedings of the International Conference on Learning Representations ({ICLR})}} }

@String { ICML         = {{Proceedings of the International Conference on Machine Learning ({ICML})}} }

@String { IJCV         = {{International Journal of Computer Vision ({IJCV})}} }

@String { NeurIPS      = {{Neural Information Processing Systems ({NeurIPS})}} }

@String { TIP          = {{{IEEE} Transactions on Image Processing}} }

@String { CVPR         = {{{IEEE} Conference on Computer Vision and Pattern Recognition ({CVPR})}} }

@String { CVPRW        = {{{IEEE} Conference on Computer Vision and Pattern Recognition Workshops ({CVPRW})}} }

@String { ECCV         = {{European Conference on Computer Vision ({ECCV})}} }

@String { ECCVW        = {{European Conference on Computer Vision Workshops ({ECCVW})}} }

@String { ICCV         = {{International Conference on Computer Vision ({ICCV})}} }

@String { ICCVW        = {{International Conference on Computer Vision Workshops ({ICCVW})}} }

@String { ICLR         = {{International Conference on Learning Representations ({ICLR})}} }

@String { ICML         = {{International Conference on Machine Learning ({ICML})}} }

@inproceedings{zhang2018unreasonable,
  title={The unreasonable effectiveness of deep features as a perceptual metric},
  author={Zhang, Richard and Isola, Phillip and Efros, Alexei A and Shechtman, Eli and Wang, Oliver},
  booktitle=CVPR,
  year={2018}
}

@inproceedings{liang2021swinir,
  title={{SwinIR}: Image restoration using swin transformer},
  author={Liang, Jingyun and Cao, Jiezhang and Sun, Guolei and Zhang, Kai and Van Gool, Luc and Timofte, Radu},
  booktitle=ICCVW,
  year={2021}
}

@inproceedings{sun2024pisasr,
  title={Pixel-level and Semantic-level Adjustable Super-resolution: A Dual-{LoRA} Approach},
  author={Sun, Lingchen and Wu, Rongyuan and Ma, Zhiyuan and Liu, Shuaizheng and Yi, Qiaosi and Zhang, Lei},
  booktitle=CVPR,
  year={2025}
}

@inproceedings{
wu2024qalign,
title={{Q-Align}: Teaching {LMM}s for Visual Scoring via Discrete Text-Defined Levels},
author={Wu, Haoning and Zhang, Zicheng and Zhang, Weixia and Chen, Chaofeng and Liao, Liang and Li, Chunyi and Gao, Yixuan and Wang, Annan and Zhang, Erli and Sun, Wenxiu and Yan, Qiong and Min, Xiongkuo and Zhai, Guangtao and Lin, Weisi},
booktitle=ICML,
year={2024},
}

@inproceedings{wu2024one,
  title={One-step effective diffusion network for real-world image super-resolution},
  author={Wu, Rongyuan and Sun, Lingchen and Ma, Zhiyuan and Zhang, Lei},
  booktitle=NeurIPS,
  year={2024}
}

@inproceedings{wang2023exploring,
  title={Exploring {CLIP} for assessing the look and feel of images},
  author={Wang, Jianyi and Chan, Kelvin CK and Loy, Chen Change},
  booktitle=AAAI, 
  year={2023}
}

@inproceedings{ke2021musiq,
  title={{MUSIQ}: Multi-scale image quality transformer},
  author={Ke, Junjie and Wang, Qifei and Wang, Yilin and Milanfar, Peyman and Yang, Feng},
  booktitle=ICCV,
  year={2021}
}

@misc{pyiqa,
  title={{IQA-PyTorch}: PyTorch Toolbox for Image Quality Assessment},
  author={Chaofeng Chen and Jiadi Mo},
  year={2022},
  howpublished = "[Online]. Available: \url{https://github.com/chaofengc/IQA-PyTorch}"
}

@article{wang2004image,
  title={Image quality assessment: from error visibility to structural similarity},
  author={Wang, Zhou and Bovik, Alan C and Sheikh, Hamid R and Simoncelli, Eero P},
  journal=TIP,
  year={2004},
}

@article{wang2024exploiting,
  title={Exploiting diffusion prior for real-world image super-resolution},
  author={Wang, Jianyi and Yue, Zongsheng and Zhou, Shangchen and Chan, Kelvin CK and Loy, Chen Change},
  journal=IJCV,
  year={2024},
}

@inproceedings{wu2024seesr,
  title={{SeeSR}: Towards semantics-aware real-world image super-resolution},
  author={Wu, Rongyuan and Yang, Tao and Sun, Lingchen and Zhang, Zhengqiang and Li, Shuai and Zhang, Lei},
  booktitle=CVPR,
  year={2024}
}

@inproceedings{yang2024pasd,
  title={Pixel-aware {Stable Diffusion} for realistic image super-resolution and personalized stylization},
  author={Yang, Tao and Wu, Rongyuan and Ren, Peiran and Xie, Xuansong and Zhang, Lei},
  booktitle=ECCV,
  year={2024},
}

@article{moser2024diffusion,
  title={Diffusion models, image super-resolution, and everything: A survey},
  author={Moser, Brian B and Shanbhag, Arundhati S and Raue, Federico and Frolov, Stanislav and Palacio, Sebastian and Dengel, Andreas},
  journal={IEEE Transactions on Neural Networks and Learning Systems},
  year={2024},
  publisher={IEEE}
}

@inproceedings{yu2024scaling,
  title={Scaling up to excellence: Practicing model scaling for photo-realistic image restoration in the wild},
  author={Yu, Fanghua and Gu, Jinjin and Li, Zheyuan and Hu, Jinfan and Kong, Xiangtao and Wang, Xintao and He, Jingwen and Qiao, Yu and Dong, Chao},
  booktitle=CVPR,
  year={2024}
}

@inproceedings{chen2025faithdiff,
  title={{FaithDiff: Unleashing diffusion priors for faithful image super-resolution}},
  author={Chen, Junyang and Pan, Jinshan and Dong, Jiangxin},
  booktitle=CVPR,
  year={2025}
}

@inproceedings{duan2025dit4sr,
  title={{DiT4SR}: Taming diffusion transformer for real-world image super-resolution},
  author={Duan, Zheng-Peng and Zhang, Jiawei and Jin, Xin and Zhang, Ziheng and Xiong, Zheng and Zou, Dongqing and Ren, Jimmy S and Guo, Chunle and Li, Chongyi},
  booktitle=ICCV,
  year={2025}
}

@inproceedings{zhang2024diffusion,
  title={Diffusion-based Blind Text Image Super-Resolution},
  author={Zhang, Yuzhe and Zhang, Jiawei and Li, Hao and Wang, Zhouxia and Hou, Luwei and Zou, Dongqing and Bian, Liheng},
  booktitle=CVPR,
  year={2024}
}

@inproceedings{aithal2025understanding,
  title={Understanding hallucinations in diffusion models through mode interpolation},
  author={Aithal, Sumukh K and Maini, Pratyush and Lipton, Zachary and Kolter, J Zico},
  booktitle=NEURIPS,
  year={2024}
}

@inproceedings{conde2022swin2sr,
  title={{Swin2SR}: {SwinV2} transformer for compressed image super-resolution and restoration},
  author={Conde, Marcos V and Choi, Ui-Jin and Burchi, Maxime and Timofte, Radu},
  booktitle=ECCVW,
  year={2022},
}

@inproceedings{lin2024diffbir,
  title={{DiffBIR}: Toward blind image restoration with generative diffusion prior},
  author={Lin, Xinqi and He, Jingwen and Chen, Ziyan and Lyu, Zhaoyang and Dai, Bo and Yu, Fanghua and Qiao, Yu and Ouyang, Wanli and Dong, Chao},
  booktitle=ECCV,
  year={2024},
}

@inproceedings{noroozi2024you,
  title={You only need one step: Fast super-resolution with stable diffusion via scale distillation},
  author={Noroozi, Mehdi and Hadji, Isma and Martinez, Brais and Bulat, Adrian and Tzimiropoulos, Georgios},
  booktitle=ECCV,
  year={2024},
}

@inproceedings{cai2019toward,
  title={Toward real-world single image super-resolution: A new benchmark and a new model},
  author={Cai, Jianrui and Zeng, Hui and Yong, Hongwei and Cao, Zisheng and Zhang, Lei},
  booktitle=ICCV,
  year={2019}
}

@inproceedings{wei2020component,
  title={Component divide-and-conquer for real-world image super-resolution},
  author={Wei, Pengxu and Xie, Ziwei and Lu, Hannan and Zhan, Zongyuan and Ye, Qixiang and Zuo, Wangmeng and Lin, Liang},
  booktitle=ECCV,
  year={2020},
}

@misc{ren2025hallucination,
  title={{Hallucination Score: Towards Mitigating Hallucinations in Generative Image Super-Resolution}},
  author={Ren, Weiming and Goyal, Raghav and Hu, Zhiming and Aumentado-Armstrong, Tristan Ty and Mohomed, Iqbal and Levinshtein, Alex},
  eprint="2507.14367",
  archivePrefix="arXiv",
  year={2025}
}

@inproceedings{min2025tair,
  title={Text-Aware Image Restoration with Diffusion Models},
  author={Min, Jaewon and Kim, Jin Hyeon and Cho, Paul Hyunbin and Lee, Jaeeun  and Park, Jihye and Park, Hyunhee  and Kim, Seungryong  and Kyu, Park Min and  Kim, Sangpil},
  booktitle=ICLR,
  year={2026}
}

@misc{kim2025unified,
  title={Unified Diffusion Transformer for High-fidelity Text-Aware Image Restoration},
  author={Kim, Jin Hyeon and Cho, Paul Hyunbin and Kim, Claire and Min, Jaewon and Lee, Jaeeun and Park, Jihye and Choi, Yeji and Kim, Seungryong},
  year={2025},
  eprint={2512.08922},
  archivePrefix={arXiv},
}

@inproceedings{hu2025text,
  title={Text-Aware Real-World Image Super-Resolution via Diffusion Model with Joint Segmentation Decoders},
  author={Hu, Qiming and Fan, Linlong and Luo, Yiyan and Yu, Yuhang and Guo, Xiaojie and Fan, Qingnan},
  booktitle=NeurIPS,
  year={2025}
}

@misc{luo2025restore,
  title={Restore Text First, Enhance Image Later: Two-Stage Scene Text Image Super-Resolution with Glyph Structure Guidance},
  author={Luo, Minxing and Fan, Linlong and Qiushi, Wang and Wu, Ge and Luo, Yiyan and Yu, Yuhang and Chen, Jinwei and Wang, Yaxing and Fan, Qingnan and Yang, Jian},
  archivePrefix="arXiv",
  eprint="2510.21590",
  year={2025}
}

@inproceedings{singh2024dcdm,
  title={{DCDM}: Diffusion-conditioned-diffusion model for scene text image super-resolution},
  author={Singh, Shrey and Keserwani, Prateek and Iwamura, Masakazu and Roy, Partha Pratim},
  booktitle=ECCV,
  pages={303--320},
  year={2024}
}

@inproceedings{mou2020plugnet,
  title={Plugnet: Degradation aware scene text recognition supervised by a pluggable super-resolution unit},
  author={Mou, Yongqiang and Tan, Lei and Yang, Hui and Chen, Jingying and Liu, Leyuan and Yan, Rui and Huang, Yaohong},
  booktitle=ECCV,
  pages={158--174},
  year={2020}
}

@inproceedings{chen2021scene,
  title={Scene text telescope: Text-focused scene image super-resolution},
  author={Chen, Jingye and Li, Bin and Xue, Xiangyang},
  booktitle=CVPR,
  pages={12026--12035},
  year={2021}
}

@InProceedings{Zhang_2022_CVPR,
    author    = {Zhang, Xiang and Su, Yongwen and Tripathi, Subarna and Tu, Zhuowen},
    title     = {Text Spotting Transformers},
    booktitle = {Proceedings of the IEEE/CVF Conference on Computer Vision and Pattern Recognition (CVPR)},
    month     = {June},
    year      = {2022},
    pages     = {9519-9528}
}

@InProceedings{Agustsson_2017_CVPR_Workshops,
	author = {Agustsson, Eirikur and Timofte, Radu},
	title = {NTIRE 2017 Challenge on Single Image Super-Resolution: Dataset and Study},
	booktitle = CVPRW,
	year = {2017}
}

@misc{yuliang2017detecting,
  title={Detecting curve text in the wild: New dataset and new solution},
  author={Yuliang, Liu and Lianwen, Jin and Shuaitao, Zhang and Sheng, Zhang},
  archivePrefix="arXiv",
  eprint="1712.02170",
  year={2017}
}

@inproceedings{ch2017total,
  title={Total-text: A comprehensive dataset for scene text detection and recognition},
  author={Ch'Ng, Chee Kheng and Chan, Chee Seng},
  booktitle={IAPR International Conference on Document Analysis and Recognition (ICDAR)},
  pages={935--942},
  year={2017}
}

@InProceedings{wang2021realesrgan,
    author    = {Xintao Wang and Liangbin Xie and Chao Dong and Ying Shan},
    title     = {{Real-ESRGAN: Training Real-World Blind Super-Resolution with Pure Synthetic Data}},
    booktitle = {International Conference on Computer Vision Workshops (ICCVW)},
    year      = {2021}
}

@misc{bai2025qwen25vltechnicalreport,
      title={Qwen2.5-VL Technical Report}, 
      author={Shuai Bai and Keqin Chen and Xuejing Liu and Jialin Wang and Wenbin Ge and Sibo Song and Kai Dang and Peng Wang and Shijie Wang and Jun Tang and Humen Zhong and Yuanzhi Zhu and Mingkun Yang and Zhaohai Li and Jianqiang Wan and Pengfei Wang and Wei Ding and Zheren Fu and Yiheng Xu and Jiabo Ye and Xi Zhang and Tianbao Xie and Zesen Cheng and Hang Zhang and Zhibo Yang and Haiyang Xu and Junyang Lin},
      year={2025},
      eprint="2502.13923",
      archivePrefix="arXiv"
}

@inproceedings{wang2020scene,
  title={Scene text image super-resolution in the wild},
  author={Wang, Wenjia and Xie, Enze and Liu, Xuebo and Wang, Wenhai and Liang, Ding and Shen, Chunhua and Bai, Xiang},
  booktitle=ECCV,
  pages={650--666},
  year={2020}
}

@inproceedings{ma2022text,
  title={A text attention network for spatial deformation robust scene text image super-resolution},
  author={Ma, Jianqi and Liang, Zhetong and Zhang, Lei},
  booktitle=CVPR,
  pages={5911--5920},
  year={2022}
}

@article{ma2023text,
  title={Text prior guided scene text image super-resolution},
  author={Ma, Jianqi and Guo, Shi and Zhang, Lei},
  journal=TIP,
  volume={32},
  pages={1341--1353},
  year={2023},
  publisher={IEEE}
}

@inproceedings{lin2017focal,
  title={Focal loss for dense object detection},
  author={Lin, Tsung-Yi and Goyal, Priya and Girshick, Ross and He, Kaiming and Doll{\'a}r, Piotr},
  booktitle=ICCV,
  pages={2980--2988},
  year={2017}
}

@inproceedings{rezatofighi2019generalized,
  title={Generalized intersection over union: A metric and a loss for bounding box regression},
  author={Rezatofighi, Hamid and Tsoi, Nathan and Gwak, JunYoung and Sadeghian, Amir and Reid, Ian and Savarese, Silvio},
  booktitle=CVPR,
  pages={658--666},
  year={2019}
}

@misc{loshchilov2017decoupled,
  title={Decoupled weight decay regularization},
  author={Loshchilov, Ilya and Hutter, Frank},
  eprint="1711.05101",
  archivePrefix="arXiv",
  year={2017}
}

@inproceedings{jie2022DASR,
  title={Efficient and Degradation-Adaptive Network for Real-World Image Super-Resolution},
  author={Liang, Jie and Zeng, Hui and Zhang, Lei},
  booktitle=ECCV,
  year={2022}
}

@misc{gemma4-31b,
  title = {{Gemma4-31B}},
  author={Google DeepMind},
  year = {2026},
  howpublished = {\url{https://ai.google.dev/gemma/docs/core/model_card_4}},
}

\clearpage
\newpage
\appendix
\clearpage
{
\begin{center}
\large{\textbf{TGSR: Text-Grounded Diffusion for Faithful Image Super-Resolution}} \\
\vspace{-5pt}

\begin{center}
\large{Supplementary Material}
\end{center}
\end{center}
}
\setcounter{section}{0}
\renewcommand\thesection{\Alph{section}}

\setcounter{page}{1}

\section{Overview}
In this supplement, we provide additional details on the hallucinated text detection dataset, including dataset statistics (\cref{sec:htd_dataset_supp}). We then describe the training and inference setup used in our experiments (\cref{sec:exp_setup_supp}). Furthermore, we present additional experimental results, including quantitative results for test-time adaptation (\cref{sec:result_tta_supp}) and the hallucinated text detector (\cref{sec:results_htd_supp}), and more qualitative examples (\cref{sec:quali_supp}). 

\section{Hallucinated Text Detection Dataset}
\label{sec:htd_dataset_supp}

\subsection{Statistics on the Hallucinated Text Instances} \label{sec:htd_stats_supp}

By varying both degradation pipelines and SR models, a single HR image produces 12 SR outputs. We take PASD \cite{yang2024pasd} as an illustrative example. When applying PASD to LR images degraded using the Real-ESRGAN configuration adopted in PASD, we obtain 105,330 SR images and 271,226 cropped text instances. After annotating with our pipeline, 63,168 (23.29\%) text regions are labelled as normal (\ie, correctly recognized), while 208,058 (76.71\%) are labelled as hallucinated. 

In contrast, when PASD is applied to LR images degraded using the Real-ESRGAN configuration adopted in StableSR \cite{wang2024exploiting}, only 38,056 (14.03\%) text regions are labelled as normal. This discrepancy arises because the PASD checkpoint is trained on LR images generated with its own degradation setting, making it better matched to that distribution and thus more capable of restoring text under consistent conditions. We show the number of hallucinated text instances for different SR models and degradation setups in \cref{tab:hti-stats}. 

\begin{figure*}[htb]
    \centering
    \includegraphics[width=0.9\linewidth]{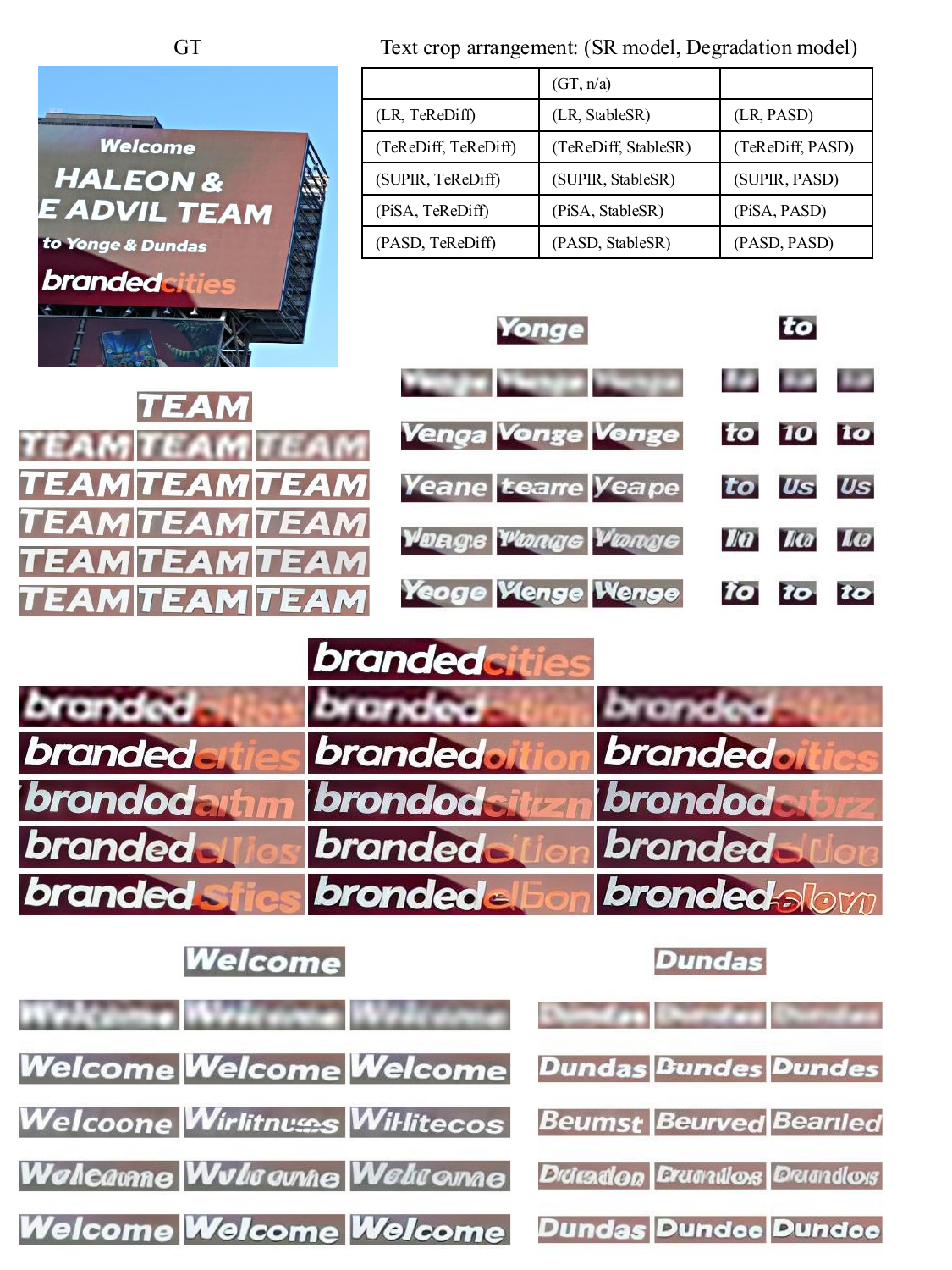}
    \caption{\textbf{Example patches from SA-Text-HTD (ours) dataset.} Most SR models recover text that is large and readable in the LR image despite the degradation (\eg, ``TEAM''). Conversely, we observe a higher amount of hallucination in the SR output of small non-readable text (\eg, ``Yonge'') and large non-readable text (\eg, ``brandedcities''). We do not visualize the remaining readable text (\ie, ``HALEON'', ``E'' and ``ADVIL'') since their SR outputs are nearly identical and non-hallucinatory.}
    \label{fig:sr_text_crops_supp}
\end{figure*}

By systematically varying degradation pipelines and SR models, our framework naturally produces diverse hallucination patterns, as shown in \cref{fig:sr_text_crops_supp}, and substantially increases the total number of hallucinated text instances, which is critical for robust training and evaluation. 
\begin{table*}[htb]
    \centering
    \begin{tabular}{l|cccc}
        \toprule
        & \multicolumn{4}{c}{SR Models} \\
        \cmidrule(lr){2-5}
        Real-ESRGAN Degradations & TeReDiff & PiSA-SR & PASD & SUPIR \\
        \midrule
        TeReDiff~\cite{min2025tair}  & 215,489 & 225,938 & 223,471 & 229,019\\
        StableSR~\cite{wang2024exploiting} & 230,040 & 235,771 & 233,170 & 238,013 \\
        PASD~\cite{yang2024pasd}     & 199,038 & 211,329 & 208,058 & 214,275\\
    \bottomrule
    \end{tabular}
    \caption{\textbf{Number of hallucinated text instances.} Each HR image corresponds to 12 SR images (three degradation levels and four SR models). This table reports the number of hallucinated text regions for each combination of degradation level and SR model. Overall, among the 271,226 total text crops, the majority correspond to the hallucinated text regions. Across degradation levels, fewer hallucinated text regions are produced when images are degraded using the PASD~\cite{yang2024pasd} degradation, likely because the SR models used in dataset curation are typically trained with degradation settings similar to those in PASD. Across models, TeReDiff~\cite{min2025tair} produces the fewest hallucinated text regions, highlighting the advantage of text-aware SR.}
    \label{tab:hti-stats}
\end{table*}

\subsection{Prompt Used for Qwen2.5-VL-72B Model}
When applying Qwen2.5-VL-72B for recognizing the texts in the text crops, we use the following prompt: \textit{Please check whether the text in the image is recognizable or not. If yes, also recognize the text in that image. Otherwise, output `N/A'. Answer only the recognized text or N/A.}

\section{Training and Inference Setup}
\label{sec:exp_setup_supp}
In this section, we present the loss functions for training the hallucinated text detector and additional implementation details. 

\subsection{Loss Functions for Hallucinated Text Detector}
In addition to the loss function for the hallucinated text prediction head, we also adopt the existing loss functions in the TESTR framework~\cite{Zhang_2022_CVPR, min2025tair} to facilitate training. Specifically, an encoder loss $\mathcal{L}_{\text{enc}}$ is applied to supervise the coarse bounding boxes generated by the transformer-based encoder module. After the decoder, which outputs fine-grained text polygons and text recognition results, an additional loss $\mathcal{L}_{\text{dec}}$ is applied to supervise these predictions. The detailed loss functions are defined as follows:
\begin{align}
\mathcal{L}_{\text{enc}} &= \sum_m \left( \lambda_{\text{cls}}\cdot \mathcal{L}^{(m)}_{\text{cls}} + \lambda_{\text{box}}\cdot \mathcal{L}^{(m)}_{\text{box}} + \lambda_{\text{gIoU}}\cdot \mathcal{L}^{(m)}_{\text{gIoU}} \right), \label{eq:ts_enc} \\
\mathcal{L}_{\text{dec}} &= \sum_n \left( \lambda_{\text{cls}}\cdot \mathcal{L}^{(n)}_{\text{cls}} + \lambda_{\text{poly}}\cdot \mathcal{L}^{(n)}_{\text{poly}} + \lambda_{\text{char}}\cdot \mathcal{L}^{(n)}_{\text{char}} \right. \nonumber \\
& \quad \left. + \mathcal{L}^{(n)}_{\mathrm{halluci\_cls}} \right ), \label{eq:ts_dec}
\end{align}
where $\lambda_\mathrm{cls}$, $\lambda_\mathrm{box}$, $\lambda_\mathrm{gIoU}$, $\lambda_\mathrm{poly}$ and $\lambda_\mathrm{char}$ are the weights for the individual loss functions. The loss weights are set as follows: $\lambda_\mathrm{cls} =2.0$, $\lambda_\mathrm{box}=5.0$, $\lambda_\mathrm{gIoU}=2.0$, $\lambda_\mathrm{poly}=5.0$ and $\lambda_\mathrm{char}=4.0$ as in TeReDiff~\cite{min2025tair}.

Instance classification utilizes the focal loss~\cite{lin2017focal} to handle class imbalance. For the $j$-th query, the classification loss is: 

\begin{align}
\mathcal{L}_{\mathrm{cls}}^{(j)} ={} & - \mathbf{1}_{\{j \in \mathrm{Im}(\sigma)\}} \, \alpha (1 - \hat{b}^{(j)})^{\gamma} \log(\hat{b}^{(j)}) \nonumber \\
& - \mathbf{1}_{\{j \notin \mathrm{Im}(\sigma)\}} \, (1 - \alpha) (\hat{b}^{(j)})^{\gamma} \log(1 - \hat{b}^{(j)}),
\end{align}
where $\hat{b}^{(\cdot)}$ is the predicted confidence score, and $\mathrm{Im}(\sigma)$ is the set of indices matched by the optimal bipartite assignment $\sigma$ between predictions and ground truth instances. $\alpha$ is class balancing parameter and $\gamma$ is the focusing parameter in focal loss~\cite{lin2017focal}. 

The control point regression loss minimizes the $\ell_1$ distance between predicted and ground truth control points:
\begin{align}
\mathcal{L}_{\mathrm{poly}}^{(j)} = \mathbf{1}_{\{j \in \mathrm{Im}(\sigma)\}} \sum_{i=1}^{N} \left\lVert \hat{d}^{(j)}_i - d^{(\sigma^{-1}(j))}_i \right\rVert_1,
\end{align}
where $\hat{d}_i^{(\cdot)}$ and $d_i^{(\cdot)}$ represent predicted and ground truth control points, respectively, and $\sigma^{-1}$ maps prediction index $j$ to its corresponding ground truth index.

Character classification is a multi-class problem and optimized via cross-entropy loss:
\begin{align}
\mathcal{L}_{\mathrm{char}}^{(j)} = \mathbf{1}_{\{j \in \mathrm{Im}(\sigma)\}} \sum_{i=1}^{M} \left( -r^{(\sigma^{-1}(j))}_i \log \hat{r}^{(j)}_i \right),
\end{align}
where $\hat{r}_i^{(\cdot)}$ denotes the predicted probability for class $i$, and $r_i^{(\cdot)}$ is the one-hot encoded ground truth label.

\subsection{Additional Implementation Details}
The hallucinated text detector is trained with a fixed timestep $ t = 0$. We use the AdamW optimizer with default parameters, a learning rate of $1 \times 10^{-4}$, a batch size of 128, and a total of 100K iterations. All other implementation details follow the default settings of TESTR~\cite{Zhang_2022_CVPR}.
For the base SR models used for our test-time adaptation experiments, we use the publicly released stage 3 checkpoint for TeReDiff~\cite{min2025tair}, the quality-optimized checkpoint with quality-oriented \texttt{SUPIR-v0Q} parameter settings for SUPIR~\cite{yu2024scaling}, and the default inference settings for PiSA-SR~\cite{sun2024pisasr} (where only the PiSA-LoRA is adopted).

\noindent
\textbf{Detector instantiation and initialization.} For each backbone, we instantiate the hallucinated text detector by freezing the backbone's diffusion U-Net and training only the text-spotting head, \ie, the transformer feature encoder, detection and recognition decoders, and the hallucination prediction head, on the multi-scale features of the frozen U-Net. For TeReDiff, we start from its pre-trained stage 3 model, whose U-Net and ControlNet already serve as the SR backbone. For PiSA-SR and SUPIR, the head is initialized from the trained TeReDiff detector's weights. An identity-initialized $1{\times}1$ adapter is added to SUPIR to bridge the gap between its SDXL-based feature space and the SD2.1-based features on which the detector head was pretrained.

\noindent
\textbf{Per-backbone guidance application.} For the perceptual term, MUSIQ scores are normalized to $[0,1]$. For TeReDiff and SUPIR, the update in \cref{eq:gradient_descent} is applied to the intermediate prediction, $\tilde{z}_0$, at every sampling step, and we normalize the gradients of $\tilde{z}_0$ before each update. PiSA-SR~\cite{sun2024pisasr}, in contrast, produces its clean latent, $z_0$, in a single step. We therefore apply a single update to $z_0$ before decoding it into the final SR image, where the mask is predicted from one additional forward pass of the same frozen U-Net on $z_0$. We apply a step size $s$ of 1, 4 and 30 for TeReDiff, SUPIR and PiSA-SR, respectively. Step sizes are not comparable across backbones: before applying $s$, we divide out the coefficient with which $\tilde{z}_0$ enters the next sample, so that a step displaces the trajectory by a uniform amount over the schedule, and this coefficient is sampler-specific (the DDPM posterior mean coefficient for TeReDiff, $\sigma_{\text{hat}}/(\sigma_{\text{hat}}-\sigma_{\text{next}})$ for SUPIR's EDM sampler). 

\subsection{Computing Infrastructure}
\label{sec:infra_supp}
Experiments were run on two machines. TeReDiff~\cite{min2025tair} and SUPIR~\cite{yu2024scaling} experiments, which dominate the compute budget, were run on eight NVIDIA H100 GPUs (80\,GB), 192 vCPUs and roughly 2\,TB of system memory, under Ubuntu 22.04.5 LTS. Each evaluation run occupies one full node and shards the test split across the eight GPUs. On TeReDiff, a complete pass over the 847 Real-Text images takes about 20 minutes for a base model and about 50 minutes with \our. PiSA-SR~\cite{sun2024pisasr} experiments were run on a single NVIDIA RTX 6000 Ada GPU (48\,GB). The per-image inference costs in \cref{tab:efficiency} are measured on a single GPU throughout, so that latency is not confounded by multi-GPU sharding.

The TeReDiff and SUPIR experiments run on Python 3.10 with PyTorch 2.4 (CUDA 12.1), \texttt{diffusers} 0.31 and \texttt{transformers} 4.57; PiSA-SR runs inside the environment released by its authors. All image quality metrics are computed with \texttt{pyiqa} 0.1.15, and the text spotter runs in a Detectron2-based environment.

\subsection{Evaluation Metric Details}
\label{sec:metrics_supp}
\noindent
\textbf{Text recognition metrics.} OCR-P, OCR-R and OCR-F1 are end-to-end text spotting metrics produced by an off-the-shelf spotter, TESTR~\cite{Zhang_2022_CVPR}. Specifically, we use the publicly released Total-Text checkpoint, and keep instances whose confidence exceeds $0.5$. We evaluate in the lexicon-free setting. OCR-P is the fraction of predicted instances that are correct, OCR-R the fraction of ground-truth instances that are correctly recovered, and OCR-F1 their harmonic mean.

We adopt these metrics because hallucinated text is, by construction, text that looks plausible but reads wrong. Full-reference image quality (FR-IQA) metrics are nearly blind to it: a hallucinated word and the correct word differ in a small fraction of the pixels of a $512{\times}512$ image and therefore barely move PSNR or SSIM. Precision is the metric of interest for our claim, as suppressing a hallucination removes an incorrect prediction (raising OCR-P) without adding a correct one, while recall measures the cost of that suppression on legible text.

\noindent
\textbf{Hallucinated text detection metrics.} For the detector we report precision and recall on the \emph{hallucinated} class only, at the deployed operating point $\tau{=}0.3$, against the hallucination labels produced by the automatic annotation pipeline of \cref{sec:autolabel}. Precision and recall are not symmetric in cost here: a false positive pulls a legible region toward the degraded reference and directly harms readability, whereas a false negative is a no-op that leaves the base model's output unchanged. We therefore select operating points by precision rather than by F1.

\noindent
\textbf{Image quality metrics.} All image quality metrics are computed with \texttt{pyiqa}~\cite{pyiqa} using its default released weights. PSNR is computed on the Y channel of the YCbCr representation; SSIM~\cite{wang2004image}, and LPIPS~\cite{zhang2018unreasonable} are full-reference metrics computed against the ground-truth HR image. MUSIQ~\cite{ke2021musiq} and CLIPIQA~\cite{wang2023exploring} are no-reference and require no ground truth; QAlign~\cite{wu2024qalign} is a no-reference multimodal quality score obtained from the \texttt{one-align} model. We report no-reference metrics alongside full-reference ones because our test-time update deliberately trades generative detail for fidelity inside hallucinated regions, and we need to verify that this does not degrade the perceptual quality of the rest of the image.

\begin{table*}[htb]
\centering
\fontsize{9}{10}\selectfont
\setlength{\tabcolsep}{5pt}
\begin{tabular}{l ccc cccc c}
\toprule
& \multicolumn{3}{c}{Inference cost} & \multicolumn{4}{c}{Image quality} & Recognition \\
\cmidrule(lr){2-4} \cmidrule(lr){5-8} \cmidrule(lr){9-9}
Method & Time (s) & $\Delta$ & Mem (GB) & PSNR$\uparrow$ & LPIPS$\downarrow$ & MUSIQ$\uparrow$ & QAlign$\uparrow$ & OCR-F1$\uparrow$ \\
\midrule
TeReDiff                        & $5.58_{\pm.09}$           & --                & \phantom{0}7.9 & 24.98 & 0.286 & 62.37 & 3.551 & 47.23 \\
\quad \emph{w/ RGB MSE only}    & $10.89_{\pm.24}$          & $+97\%$           & 14.6 & 25.27 & 0.284 & 60.93 & 3.430 & 52.19 \\
\quad +TGSR-lite & $7.55_{\pm.15}$           & $+35\%$           & \phantom{0}7.8 & 25.17 & 0.283 & 61.39 & 3.446 & 52.59 \\
\quad +TGSR            & $13.48_{\pm.26}$ & $+141\%$ & 15.7 & 25.05 & 0.286 & 67.24 & 3.534 & 52.89 \\
\midrule
PiSA-SR                         & $0.124_{\pm.004}$ & --                & \phantom{0}9.6 & 26.46 & 0.241 & 64.37 & 3.260 & 46.86 \\
\quad \emph{w/ RGB MSE only}    & $0.408_{\pm.010}$ & $+229\%$          & 14.1 & 26.51 & 0.242 & 63.72 & 3.218 & 49.92 \\
\quad +TGSR-lite & $0.243_{\pm.003}$ & $+96\%$           & \phantom{0}9.6 & 26.52 & 0.243 & 63.61 & 3.204 & 51.10 \\
\quad +TGSR            & $0.420_{\pm.004}$ & $+239\%$ & 15.8 & 26.52 & 0.244 & 64.72 & 3.212 & 50.98 \\
\midrule
SUPIR                           & $16.13_{\pm.96}$           & --                & 26.2 & 24.16 & 0.394 & 59.58 & 3.481 & 40.32 \\
\quad \emph{w/ RGB MSE only}    & $32.30_{\pm1.02}$          & $+97\%$           & 45.5 & 24.29 & 0.393 & 59.14 & 3.441 & 41.93 \\
\quad +TGSR-lite & $18.56_{\pm.98}$           & $+16\%$           & 26.6 & 24.30 & 0.392 & 59.11 & 3.425 & 43.02 \\
\quad +TGSR            & $35.30_{\pm1.09}$ & $+119\%$ & 48.3 & 24.27 & 0.392 & 60.81 & 3.463 & 42.58 \\
\bottomrule
\end{tabular}
\caption{\textbf{Inference cost of different models.} \textbf{Time} is the mean wall-clock
time to restore one image and \textbf{Mem} is the peak GPU memory that restoration allocates,
both measured on a single GPU with images processed one at a time; subscripts are the
standard deviation across images. $\Delta$ is the extra time over the base model
of the same block, \emph{w/ RGB MSE only} is an alternative to our default, and \our-lite is the variant that keeps only the masked latent
MSE term of the guidance objective. PiSA-SR was profiled on one
NVIDIA RTX 6000 Ada (48\,GB) over all 847 Real-Text images, whereas TeReDiff and SUPIR were
profiled on one NVIDIA H100 (80\,GB) over 100 images.}
\label{tab:efficiency}
\end{table*}

\section{Additional Results on Test-Time Adaptation}
\label{sec:result_tta_supp}
{
\setlength{\tabcolsep}{0.01in}
\fontsize{9}{10}\selectfont
\aboverulesep = 0mm
\belowrulesep = 0mm
\begin{table*}[t]
    \centering
    \sisetup{
        detect-weight,
        mode=text,
        input-symbols={},
        table-format=2.3,
        text-series-to-math = true,
        propagate-math-font = true,
    }
    \begin{adjustbox}{max width=\linewidth}
    \begin{tabular}{@{}lS[table-format=2.2]S[table-format=1.3]S[table-format=1.3]S[table-format=2.2]S[table-format=1.3]S[table-format=1.3]S[table-format=2.2]S[table-format=2.2]S[table-format=2.2]@{}}
    \toprule
      & {PSNR $\uparrow$} & {SSIM $\uparrow$} & {LPIPS $\downarrow$} & {MUSIQ $\uparrow$} & {CLIPIQA $\uparrow$} & {QAlign $\uparrow$} & {OCR-P $\uparrow$} & {OCR-R $\uparrow$ }& {OCR-F1 $\uparrow$ } \\
    \midrule
      Bicubic & 27.86 & 0.858 & 0.362 & 24.13 & 0.283 & 1.621 & 68.84 & 22.13 & 33.50 \\
      Swin2SR \cite{conde2022swin2sr} & 27.64 & 0.879 & 0.248 & 48.43 & 0.367 & 2.509 & 63.87 & 30.75 & 41.51 \\
      ESRGAN \cite{wang2021realesrgan} & 25.69 & 0.846 & 0.231 & 58.58 & 0.580 & 3.228 & 58.85 & 30.43 & 40.11 \\
      StableSR~\cite{wang2024exploiting} & 26.44 & 0.826 & 0.264 & 57.12 & 0.592 & 3.026 & 58.10 & 36.30 & 44.68 \\
      PASD~\cite{yang2024pasd} & 26.15 & 0.831 & 0.265 & 57.01 & 0.528 & 3.208 & 59.81 & 39.90 & 47.87 \\
      SeeSR~\cite{wu2024seesr} & 25.69 & 0.829 & 0.279 & 67.09 & 0.654 & 3.667 & 54.94 & 37.75 & 44.75 \\
      FaithDiff~(\citeauthor{chen2025faithdiff}) & 25.75 & 0.795 & 0.297 & 64.94 & 0.604 & 3.486 & 56.88 & 39.85 & 46.87 \\
      DiT4SR~\cite{duan2025dit4sr} & 24.36 & 0.782 & 0.296 & 63.88 & 0.625 & 3.397 & 53.99 & 43.35 & 48.09 \\
     \cdashline{1-10}\noalign{\vskip 2pt}
      PiSA-SR~\cite{sun2024pisasr} & 26.46 & 0.851 & \textbf{0.241} & 64.37 & 0.611 & \textbf{3.260} & 54.71 & \textbf{40.98} & 46.86 \\
      
      \textbf{PiSA-SR +TGSR-lite (ours)} & \textbf{26.52} & \textbf{0.852} & 0.243 & 63.61 & 0.608 & 3.204 & \textbf{68.44} & 40.76 & \textbf{51.10} \\

      \textbf{PiSA-SR +TGSR (ours)} & \textbf{26.52} & \textbf{0.852} & 0.244 & \textbf{64.72} & \textbf{0.615} & 3.212 & 68.33 & 40.66 & 50.98 \\
      \cdashline{1-10}\noalign{\vskip 2pt}
      SUPIR~\cite{yu2024scaling} & 24.16 & 0.706 & 0.394 & 59.58 & 0.659 & \textbf{3.481} & 44.94 & \textbf{36.56} & 40.32 \\
      
      \textbf{SUPIR +TGSR-lite (ours)} & \textbf{24.30} & \textbf{0.710} & \textbf{0.392} & 59.11 & 0.651 & 3.425 & \textbf{52.69} & 36.35 & \textbf{43.02} \\

      \textbf{SUPIR +TGSR (ours)} & 24.27 & 0.709 & \textbf{0.392} & \textbf{60.81} & \textbf{0.663} & 3.463 & 51.29 & 36.40 & 42.58 \\
      \cdashline{1-10}\noalign{\vskip 2pt}
      TeReDiff~\cite{min2025tair} & 24.98 & 0.809 & 0.286 & 62.37 & 0.643 & \textbf{3.551} & 53.24 & \textbf{42.43} & 47.23 \\
     
      \textbf{TeReDiff +TGSR-lite (ours)} & \textbf{25.17} & \textbf{0.813} & \textbf{0.283} & 61.39 & 0.631 & 3.446 & \textbf{70.93} & 41.79 & 52.59 \\

      \textbf{TeReDiff +TGSR (ours)} & 25.05 & 0.811 & 0.286 & \textbf{67.24} & \textbf{0.697} & 3.534 & 70.92 & 42.16 & \textbf{52.89} \\

    \bottomrule
    \end{tabular}
    \end{adjustbox}
    \caption{
    \textbf{SR results on the Real-Text dataset~\cite{min2025tair}}. Metrics and layout follow \cref{tab:real-text-main}. We compare base SR
    models against our adapted models with test-time adaptation, denoted
    \textbf{+TGSR (ours)}; \textbf{+\our-lite} is the variant that keeps only the masked latent MSE term of the guidance objective. \textbf{bold} marks the better value within each backbone. Across all three backbones, \our raises recognition precision (OCR-P) and OCR-F1 while
    preserving perceptual quality.}
    \label{tab:real-text-supp}
\end{table*}
}

\begin{figure*}[tb]
  \centering
  \begin{minipage}{0.9\linewidth}
    \includegraphics[width=\linewidth]{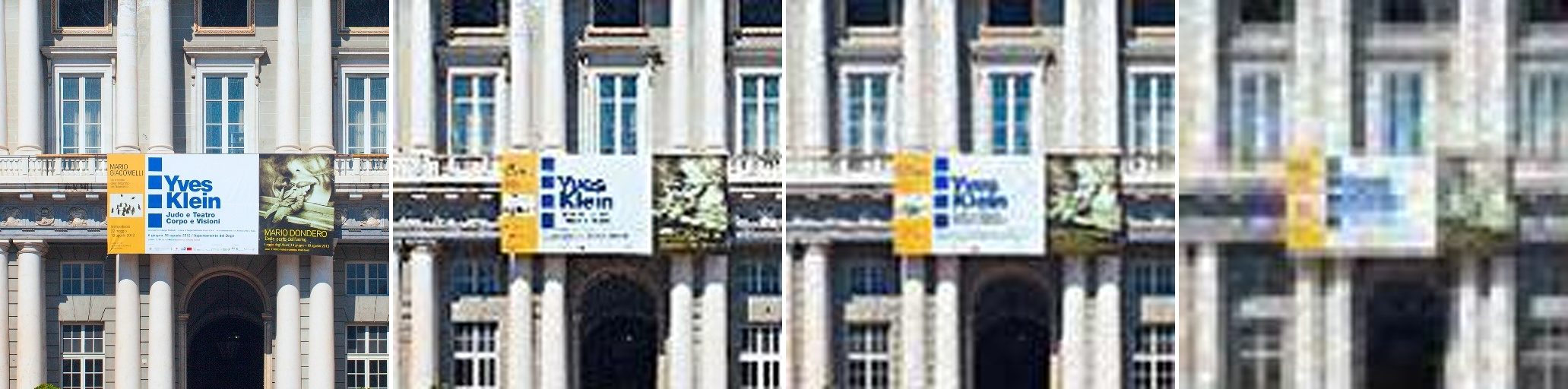}
  \end{minipage}
  \vfill
  \begin{minipage}{0.9\linewidth}
    \includegraphics[width=\linewidth]{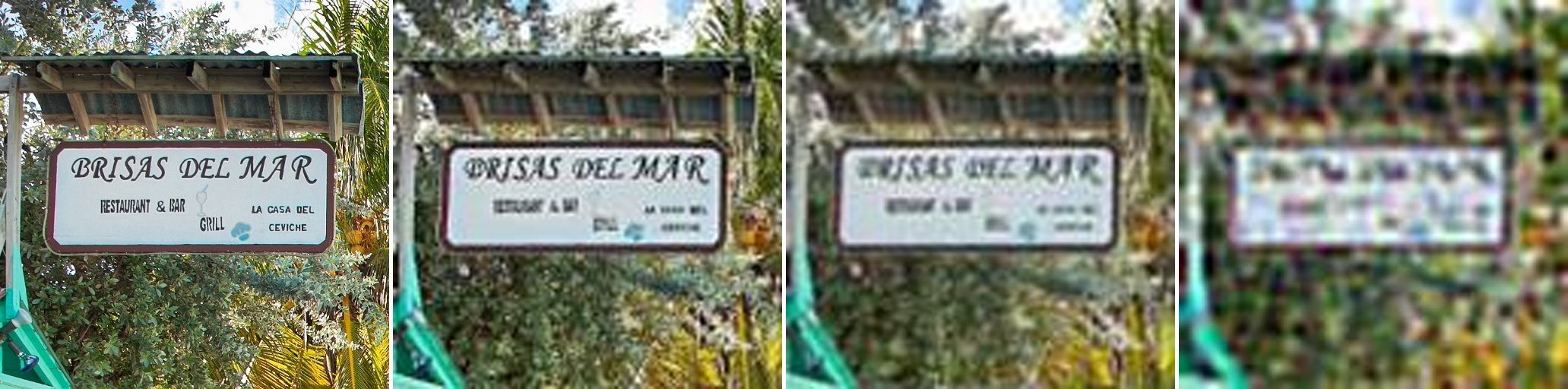}
    \vfill \vspace*{-0.12cm}
    \text{ ~~~~~~~~~~~~~~~~~~~GT~~~~~~~~~~~~~~~~~~~~~~~~~~~~~~~~~~~~Level 1~~~~~~~~~~~~~~~~~~~~~~~~~~~~~~~~~~Level 2~~~~~~~~~~~~~~~~~~~~~~~~~~~~~~~~~Level 3}
    \vfill \vspace*{0.12cm}
  \end{minipage}
  \caption{\textbf{Illustration of three levels of degradations in SA-Text-Test dataset}. The task of producing faithful super-resolution for the text region becomes increasingly difficult with higher levels of degradation.
  }
  \label{fig:3_deg_vis}
\end{figure*}

\subsection{Results on Real-Text with \our-lite}
\label{sec:realtext_lite_supp}
\cref{tab:real-text-supp} reproduces the main-paper Real-Text comparison
(\cref{tab:real-text-main}) in full and adds an additional configuration of
\textbf{+TGSR-lite}. \our-lite keeps only the masked latent MSE term of the guidance objective and drops the MUSIQ and CLIPIQA terms. The NR-IQA objectives are defined for RGB images only, so the full objective has to decode the intermediate SR output inside the guidance loop and backpropagate through the decoder to return the gradient to the latent. On the contrary, \our-lite compares latents directly, which greatly reduces the runtime
overhead. \cref{tab:efficiency} reports the resulting inference time and memory. Overall, \our-lite remains competitive in the OCR performance, with a small degradation in no-reference perceptual quality compared to \our. Specifically, \our-lite attains the higher OCR-F1 on two of the three backbones ($51.10$ \vs $50.98$ on PiSA-SR and $43.02$ \vs $42.58$ on SUPIR) and trails by only $0.30$ on TeReDiff, confirming that the recognition gain is delivered by the masked latent MSE term alone. The perceptual terms are instead what keeps
no-reference quality from regressing: \our-lite falls below the base model on both MUSIQ and CLIPIQA on all three backbones, whereas \our exceeds it on both, leading \our-lite by $1.11$, $1.70$ and $5.85$ on MUSIQ. Since MUSIQ and CLIPIQA both enter the guidance objective, the more informative comparison is QAlign. While both variants remain marginally below the base model on QAlign, \our is the closer of the two throughout, suggesting that the perceptual terms improve
no-reference quality beyond the two metrics they directly optimize.

{
\setlength{\tabcolsep}{0.01in}
\fontsize{9}{10}\selectfont
\begin{table*}[hb]
    \centering
    \sisetup{
        detect-weight,
        mode=text,
        input-symbols={},
        table-format=2.3,
        text-series-to-math = true,
        propagate-math-font = true,
    }
    \begin{adjustbox}{max width=\linewidth}
    \begin{tabular}{@{}llS[table-format=2.2]S[table-format=1.3]S[table-format=1.3]S[table-format=2.2]S[table-format=1.3]S[table-format=1.3]S[table-format=2.2]S[table-format=2.2]S[table-format=2.2]@{}}
    \toprule
      & & {PSNR $\uparrow$} & {SSIM $\uparrow$} & {LPIPS $\downarrow$} & {MUSIQ $\uparrow$} & {CLIPIQA $\uparrow$} & {QAlign $\uparrow$} & {OCR-P $\uparrow$} & {OCR-R $\uparrow$} & {OCR-F1 $\uparrow$} \\
    \midrule
      \multirow{17}{*}{Level 1}
      & Bicubic & 23.99 & 0.681 & 0.571 & 29.85 & 0.492 & 1.638 & 61.37 & 16.64 & 26.18 \\
      & Swin2SR~\cite{conde2022swin2sr} & 24.09 & 0.729 & 0.345 & 57.92 & 0.328 & 3.020 & 53.59 & 16.91 & 25.71 \\
      & Real-ESRGAN~\cite{wang2021realesrgan} & 22.65 & 0.690 & 0.235 & 69.76 & 0.567 & 3.857 & 45.40 & 16.32 & 24.01 \\
      & StableSR~\cite{wang2024exploiting} & 22.08 & 0.646 & 0.235 & 71.87 & 0.680 & 3.972 & 38.65 & 18.45 & 24.98 \\
      & PASD~\cite{yang2024pasd} & 22.61 & 0.653 & 0.264 & 72.06 & 0.649 & 4.211 & 39.71 & 20.81 & 27.31 \\
      & SeeSR~\cite{wu2024seesr} & 22.71 & 0.669 & 0.232 & 73.85 & 0.692 & 4.493 & 35.54 & 19.27 & 24.99 \\
      & FaithDiff~\cite{chen2025faithdiff} & 22.82 & 0.655 & 0.222 & 73.76 & 0.650 & 4.244 & 39.50 & 22.17 & 28.40 \\
      & DiT4SR~\cite{duan2025dit4sr} & 21.14 & 0.619 & 0.267 & 73.01 & 0.662 & 4.049 & 40.47 & 24.16 & 30.26 \\
      \cdashline{2-11}\noalign{\vskip 3pt}
      & PiSA-SR~\cite{sun2024pisasr} & 22.89 & 0.679 & \textbf{0.207} & 73.70 & 0.683 & \textbf{4.294} & 39.30 & \textbf{21.89} & 28.12 \\
      & \textbf{PiSA-SR +TGSR-lite (ours)} & \textbf{22.95} & \textbf{0.680} & 0.209 & 73.07 & 0.680 & 4.187 & 52.34 & 20.76 & 29.73 \\
      & \textbf{PiSA-SR +TGSR (ours)} & \textbf{22.95} & 0.679 & 0.209 & \textbf{74.61} & \textbf{0.721} & 4.197 & \textbf{52.99} & 20.90 & \textbf{29.97} \\
      \cdashline{2-11}\noalign{\vskip 3pt}
      & SUPIR~\cite{yu2024scaling} & 21.39 & 0.576 & \textbf{0.310} & 71.77 & 0.749 & \textbf{4.407} & 29.97 & \textbf{21.53} & 25.06 \\
      & \textbf{SUPIR +TGSR-lite (ours)} & \textbf{21.52} & \textbf{0.579} & 0.312 & 71.32 & 0.740 & 4.277 & 38.31 & 20.49 & 26.70 \\
      & \textbf{SUPIR +TGSR (ours)} & 21.50 & 0.578 & 0.312 & \textbf{72.22} & \textbf{0.755} & 4.312 & \textbf{38.43} & 20.63 & \textbf{26.84} \\
      \cdashline{2-11}\noalign{\vskip 3pt}
      & TeReDiff~\cite{min2025tair} & 22.45 & 0.661 & \textbf{0.214} & 72.23 & 0.694 & \textbf{4.360} & 37.14 & \textbf{22.12} & 27.73 \\
      & \textbf{TeReDiff +TGSR-lite (ours)} & \textbf{22.64} & \textbf{0.665} & 0.216 & 71.08 & 0.675 & 4.200 & \textbf{54.43} & 21.17 & 30.48 \\
      & \textbf{TeReDiff +TGSR (ours)} & 22.55 & 0.663 & 0.217 & \textbf{74.93} & \textbf{0.753} & 4.277 & 54.15 & 21.31 & \textbf{30.58} \\
    \midrule
      \multirow{17}{*}{Level 2}
      & Bicubic & 23.75 & 0.642 & 0.675 & 24.66 & 0.407 & 1.370 & 54.90 & 12.19 & 19.96 \\
      & Swin2SR~\cite{conde2022swin2sr} & 24.11 & 0.708 & 0.387 & 55.09 & 0.322 & 2.880 & 49.32 & 13.24 & 20.87 \\
      & Real-ESRGAN~\cite{wang2021realesrgan} & 22.73 & 0.676 & 0.257 & 70.13 & 0.577 & 3.901 & 39.81 & 13.19 & 19.82 \\
      & StableSR~\cite{wang2024exploiting} & 22.09 & 0.621 & 0.263 & 71.50 & 0.685 & 3.917 & 34.25 & 15.87 & 21.69 \\
      & PASD~\cite{yang2024pasd} & 22.44 & 0.635 & 0.295 & 71.70 & 0.642 & 4.176 & 35.05 & 17.86 & 23.66 \\
      & SeeSR~\cite{wu2024seesr} & 22.62 & 0.652 & 0.257 & 73.85 & 0.688 & 4.464 & 32.58 & 17.00 & 22.34 \\
      & FaithDiff~\cite{chen2025faithdiff} & 22.63 & 0.633 & 0.253 & 73.85 & 0.649 & 4.213 & 36.68 & 19.72 & 25.65 \\
      & DiT4SR~\cite{duan2025dit4sr} & 20.98 & 0.602 & 0.291 & 73.05 & 0.657 & 4.035 & 35.64 & 21.21 & 26.60 \\
      \cdashline{2-11}\noalign{\vskip 3pt}
      & PiSA-SR~\cite{sun2024pisasr} & 23.02 & \textbf{0.667} & \textbf{0.227} & 73.36 & 0.673 & \textbf{4.265} & 34.33 & \textbf{18.22} & 23.81 \\
      & \textbf{PiSA-SR +TGSR-lite (ours)} & \textbf{23.10} & \textbf{0.667} & 0.231 & 72.56 & 0.671 & 4.127 & \textbf{48.38} & 17.54 & \textbf{25.75} \\
      & \textbf{PiSA-SR +TGSR (ours)} & 23.09 & \textbf{0.667} & 0.231 & \textbf{74.25} & \textbf{0.713} & 4.138 & 47.10 & 17.32 & 25.32 \\
      \cdashline{2-11}\noalign{\vskip 3pt}
      & SUPIR~\cite{yu2024scaling} & 21.20 & 0.550 & \textbf{0.346} & 71.36 & 0.743 & \textbf{4.366} & 24.66 & \textbf{18.09} & 20.87 \\
      & \textbf{SUPIR +TGSR-lite (ours)} & \textbf{21.33} & \textbf{0.553} & 0.347 & 70.83 & 0.733 & 4.220 & \textbf{29.75} & 17.04 & \textbf{21.67} \\
      & \textbf{SUPIR +TGSR (ours)} & 21.32 & \textbf{0.553} & 0.348 & \textbf{71.80} & \textbf{0.746} & 4.253 & 28.63 & 17.00 & 21.33 \\
      \cdashline{2-11}\noalign{\vskip 3pt}
      & TeReDiff~\cite{min2025tair} & 22.37 & 0.649 & \textbf{0.234} & 72.33 & 0.699 & \textbf{4.374} & 35.15 & \textbf{20.76} & 26.10 \\
      & \textbf{TeReDiff +TGSR-lite (ours)} & \textbf{22.58} & \textbf{0.653} & 0.237 & 71.03 & 0.678 & 4.174 & \textbf{53.42} & 19.49 & \textbf{28.56} \\
      & \textbf{TeReDiff +TGSR (ours)} & 22.49 & 0.651 & 0.237 & \textbf{74.89} & \textbf{0.754} & 4.247 & 53.10 & 19.40 & 28.42 \\
    \midrule
      \multirow{17}{*}{Level 3}
      & Bicubic & 22.34 & 0.575 & 0.801 & 18.38 & 0.310 & 1.098 & 49.21 & 7.07 & 12.37 \\
      & Swin2SR~\cite{conde2022swin2sr} & 23.15 & 0.664 & 0.464 & 46.72 & 0.294 & 2.465 & 43.26 & 7.71 & 13.08 \\
      & Real-ESRGAN~\cite{wang2021realesrgan} & 21.86 & 0.632 & 0.321 & 69.21 & 0.595 & 3.806 & 34.10 & 8.07 & 13.05 \\
      & StableSR~\cite{wang2024exploiting} & 21.20 & 0.540 & 0.369 & 67.26 & 0.662 & 3.538 & 27.09 & 10.15 & 14.77 \\
      & PASD~\cite{yang2024pasd} & 21.34 & 0.589 & 0.376 & 68.80 & 0.597 & 3.944 & 26.60 & 12.42 & 16.93 \\
      & SeeSR~\cite{wu2024seesr} & 21.38 & 0.599 & 0.319 & 73.75 & 0.688 & 4.403 & 24.45 & 12.19 & 16.27 \\
      & FaithDiff~\cite{chen2025faithdiff} & 21.30 & 0.566 & 0.327 & 73.66 & 0.645 & 4.129 & 24.98 & 12.74 & 16.87 \\
      & DiT4SR~\cite{duan2025dit4sr} & 19.89 & 0.552 & 0.348 & 73.10 & 0.655 & 3.987 & 27.04 & 13.96 & 18.42 \\
      \cdashline{2-11}\noalign{\vskip 3pt}
      & PiSA-SR~\cite{sun2024pisasr} & 22.09 & 0.615 & \textbf{0.289} & 71.83 & 0.653 & \textbf{4.048} & 27.10 & \textbf{11.56} & 16.21 \\
      & \textbf{PiSA-SR +TGSR-lite (ours)} & \textbf{22.17} & \textbf{0.616} & 0.294 & 70.79 & 0.648 & 3.875 & 40.46 & 11.24 & 17.59 \\
      & \textbf{PiSA-SR +TGSR (ours)} & 22.16 & \textbf{0.616} & 0.294 & \textbf{72.86} & \textbf{0.691} & 3.889 & \textbf{40.92} & 11.24 & \textbf{17.64} \\
      \cdashline{2-11}\noalign{\vskip 3pt}
      & SUPIR~\cite{yu2024scaling} & 20.41 & 0.494 & \textbf{0.438} & 68.04 & 0.714 & \textbf{4.089} & 18.55 & \textbf{11.24} & 14.00 \\
      & \textbf{SUPIR +TGSR-lite (ours)} & \textbf{20.53} & \textbf{0.497} & 0.439 & 67.53 & 0.703 & 3.974 & 21.45 & 10.83 & 14.40 \\
      & \textbf{SUPIR +TGSR (ours)} & 20.51 & 0.496 & 0.439 & \textbf{68.67} & \textbf{0.718} & 4.017 & \textbf{22.37} & 10.79 & \textbf{14.56} \\
      \cdashline{2-11}\noalign{\vskip 3pt}
      & TeReDiff~\cite{min2025tair} & 21.26 & 0.605 & \textbf{0.284} & 72.12 & 0.706 & \textbf{4.347} & 27.73 & \textbf{14.73} & 19.24 \\
      & \textbf{TeReDiff +TGSR-lite (ours)} & \textbf{21.46} & \textbf{0.608} & 0.289 & 70.61 & 0.684 & 4.097 & \textbf{49.12} & 13.96 & \textbf{21.74} \\
      & \textbf{TeReDiff +TGSR (ours)} & 21.39 & 0.607 & 0.289 & \textbf{74.55} & \textbf{0.756} & 4.175 & 48.43 & 14.01 & 21.73 \\
    \bottomrule
    \end{tabular}
    \end{adjustbox}
    \caption{
    \textbf{SR results on SA-Text-Test~\cite{min2025tair} under three degradation levels.}
    Metrics and layout follow \cref{tab:real-text-main}. We compare base SR
    models against our adapted models with test-time adaptation, denoted
    \textbf{+TGSR (ours)}; \textbf{+\our-lite} is the variant that keeps only the masked latent MSE term of the guidance objective. \textbf{bold} marks the better value within each block. Across all three backbones and all three degradation
    levels, \our raises recognition precision (OCR-P) and OCR-F1 while
    preserving perceptual quality, and the precision gain grows with
    degradation severity.
    }
    \label{tab:syn-text-3deg-supp}
\end{table*}
}

\subsection{Results on SA-Text-Test} Following~\cite{min2025tair}, we present the test-time adaptation results on SA-Text-Test under three levels of degradations of increasing difficulty in \cref{tab:syn-text-3deg-supp}. This allows us to evaluate model robustness under progressively more challenging conditions as shown in Fig. \ref{fig:3_deg_vis}. In general, stronger degradations increase the likelihood of hallucinated text artifacts, leading to a drop in recognition F1 (OCR-F1). Overall, \our improves the text recognition precision (OCR-P) and F1 score while preserving the image quality metrics. Notably, \our and \our-lite with TeReDiff achieve the highest OCR-F1 among all compared methods at every degradation level. The improvement in OCR-P is most pronounced (from 27.73 to 49.12) under the most severe level 3 degradation, where hallucinated text artifacts occur most frequently without sufficient visual cues in the LR image. This highlights the effectiveness of \our in reducing the number of false positive instances (\ie, text regions that have been incorrectly super-resolved), while maintaining image perceptual quality. Similar to the Real-Text benchmark, \our-lite shows OCR performance competitive with \our, yet falls below the base model on both NR-IQA metrics in every setting, trailing \our by $0.90$--$3.94$ in MUSIQ and $0.013$--$0.078$ in CLIPIQA.

{
\setlength{\tabcolsep}{0.01in}
\fontsize{9}{10}\selectfont
\begin{table*}[t]
    \centering
    \sisetup{
        detect-weight,
        mode=text,
        input-symbols={},
        table-format=2.3,
        text-series-to-math = true,
        propagate-math-font = true,
    }
    \begin{adjustbox}{max width=\linewidth}
    \begin{tabular}{@{}lS[table-format=2.2]S[table-format=1.3]S[table-format=1.3]S[table-format=2.2]S[table-format=1.3]S[table-format=1.3]S[table-format=2.2]S[table-format=2.2]S[table-format=2.2]@{}}
    \toprule
      Configuration & {PSNR $\uparrow$} & {SSIM $\uparrow$} & {LPIPS $\downarrow$} & {MUSIQ $\uparrow$} & {CLIPIQA $\uparrow$} & {QAlign $\uparrow$} & {OCR-P $\uparrow$} & {OCR-R $\uparrow$} & {OCR-F1 $\uparrow$} \\
    \midrule
      TeReDiff~\cite{min2025tair}  & 24.98 & 0.809 & 0.286 & 62.37 & 0.643 & 3.551 & 53.24 & 42.43 & 47.23 \\
      \textbf{Ours}  & 25.05 & 0.811 & 0.286 & 67.24 & 0.697 & 3.534 & 70.92 & 42.16 & 52.89 \\
    \midrule
      \multicolumn{10}{l}{\emph{Guidance space and objective}} \\
      \quad RGB, MSE & 25.27 & 0.813 & 0.284 & 60.93 & 0.627 & 3.430 & 68.60 & 42.11 & 52.19 \\
      \quad RGB, $+$\,MUSIQ & 25.09 & 0.810 & 0.288 & 69.01 & 0.654 & 3.577 & 62.01 & 42.27 & 50.27 \\
      \quad RGB, $+$\,CLIPIQA & 25.21 & 0.812 & 0.285 & 61.56 & 0.747 & 3.479 & 62.26 & 41.84 & 50.05 \\
      \quad RGB, $+$\,MUSIQ\,$+$\,CLIPIQA & 25.08 & 0.810 & 0.288 & 68.52 & 0.719 & 3.585 & 61.18 & 42.43 & 50.11 \\
      \quad Latent, MSE & 25.17 & 0.813 & 0.283 & 61.39 & 0.631 & 3.446 & 70.93 & 41.79 & 52.59 \\
      \quad Latent, $+$\,MUSIQ & 25.06 & 0.811 & 0.286 & 67.62 & 0.646 & 3.531 & 70.91 & 42.00 & 52.76 \\
      \quad Latent, $+$\,CLIPIQA & 25.14 & 0.812 & 0.284 & 61.69 & 0.724 & 3.470 & 69.66 & 41.30 & 51.86 \\
    \midrule
      \multicolumn{10}{l}{\emph{MUSIQ loss weight $\lambda_m$}} \\
      \quad $\lambda_m = 10^{-5}$ & 25.08 & 0.811 & 0.286 & 66.37 & 0.642 & 3.511 & 70.86 & 41.90 & 52.66 \\
      \quad $\lambda_m = 10^{-4}$ & 25.07 & 0.811 & 0.286 & 66.66 & 0.643 & 3.514 & 70.83 & 41.84 & 52.61 \\
      \quad $\lambda_m = 10^{-3}$ (default) & 25.06 & 0.811 & 0.286 & 67.62 & 0.646 & 3.531 & 70.91 & 42.00 & 52.76 \\
      \quad $\lambda_m = 10^{-2}$ & 24.97 & 0.809 & 0.288 & 68.98 & 0.656 & 3.590 & 64.20 & 42.11 & 50.86 \\
    \midrule
      \multicolumn{10}{l}{\emph{CLIPIQA loss weight $\lambda_c$}} \\
      \quad $\lambda_c = 10^{-5}$ & 25.14 & 0.812 & 0.284 & 61.64 & 0.709 & 3.468 & 69.64 & 41.14 & 51.73 \\
      \quad $\lambda_c = 10^{-4}$ (default) & 25.14 & 0.812 & 0.284 & 61.69 & 0.724 & 3.470 & 69.66 & 41.30 & 51.86 \\
      \quad $\lambda_c = 10^{-3}$ & 25.11 & 0.811 & 0.285 & 61.91 & 0.746 & 3.497 & 66.21 & 41.68 & 51.16 \\
      \quad $\lambda_c = 10^{-2}$ & 24.98 & 0.809 & 0.286 & 62.63 & 0.765 & 3.564 & 54.12 & 42.49 & 47.60 \\
    \midrule
      \multicolumn{10}{l}{\emph{Empty-mask step damping $\rho$}} \\
      \quad $\rho = 0$ (no update) & 25.15 & 0.812 & 0.284 & 62.80 & 0.650 & 3.469 & 70.75 & 41.68 & 52.46 \\
      \quad $\rho = 0.5$ & 24.99 & 0.810 & 0.288 & 69.05 & 0.719 & 3.570 & 69.93 & 41.95 & 52.44 \\
      \quad $\rho = 1$ & 24.83 & 0.807 & 0.291 & 71.76 & 0.759 & 3.638 & 69.61 & 41.57 & 52.06 \\
    \bottomrule
    \end{tabular}
    \end{adjustbox}
    \caption{\textbf{Ablation studies on Real-Text},
    using TeReDiff~\cite{min2025tair} as the base model. This is the extended
    version of \cref{tab:ablation_v3} in the main paper, including all IQA metrics. Our default guides a masked
    MSE objective in latent space, adds MUSIQ ($10^{-3}$) and CLIPIQA
    ($10^{-4}$) on the decoded image, and damps the update by $\rho{=}0.3$ on
    steps where the detector finds no hallucinated region.}
    \label{tab:ablation_supp}
\end{table*}
}

\subsection{Additional Ablation Studies}
\noindent

\cref{tab:ablation_supp} is the extended version of \cref{tab:ablation_v3} in the main paper, but with additional ablations and full IQA metrics reported. Each block varies a single component of our default configuration, \ie, a masked MSE objective guided in latent space, MUSIQ ($\lambda_m{=}10^{-3}$) and CLIPIQA ($\lambda_c{=}10^{-4}$) applied to the decoded image, and an update damped by $\rho{=}0.3$ on images where the detector finds no hallucinated region. 

\noindent\textbf{Guidance space.}
We compare performing guidance in RGB space versus latent space using various objectives. Under MSE alone, latent guidance improves OCR-P by 2.33 points (68.60 to 70.93) and OCR-F1 by 0.40 points (52.19 to 52.59) at comparable image quality. The gap widens once the perceptual terms are added: in RGB space, adding MUSIQ degrades OCR-P by 6.59 points (68.60 to 62.01) and OCR-F1 by 1.92 points. We therefore adopt latent-space guidance as our default.

\noindent\textbf{Guidance objective.}
The masked reconstruction term pulls hallucinated regions toward a degraded reference, so latent MSE shows slightly worse no-reference metrics (61.39 \vs 62.37 MUSIQ, 0.631 \vs 0.643 CLIPIQA). Adding MUSIQ or CLIPIQA improves the corresponding NR-IQA metric but leaves the other NR-IQA metrics essentially unchanged. Only the combination, our default, exceeds the base model on both (67.24 MUSIQ, 0.697 CLIPIQA), while attaining the best OCR-F1 in the table (52.89).

\noindent\textbf{Perceptual loss weights.}
The two loss weights $\lambda_m$ and $\lambda_c$ behave similarly. At smaller weights, the OCR performance and IQA metrics are relatively similar. Above the default weights, recognition performance degrades, primarily due to a loss of precision. At $\lambda_m{=}10^{-2}$, OCR-P falls from 70.91 to 64.20 while OCR-R rises from 42.00 to 42.11, and at $\lambda_c{=}10^{-2}$, OCR-P falls from 69.66 to 54.12 while OCR-R improves to 42.49. The reason is that no-reference quality metrics reward sharp, plausible detail, which is exactly what a hallucination looks like, so an over-weighted perceptual term reintroduces the artifacts the reconstruction term is meant to suppress. Each weight must therefore stay at or below the default used in this paper.

\noindent\textbf{Empty-mask step damping.}
$\rho$ scales the update on images where the detector finds no hallucinated region and hence only the IQA terms play a role. At $\rho{=}0$ these images are left untouched: OCR-F1 already improves to 52.46, but no-reference quality stays at the base model's level (62.80 MUSIQ and 0.650 CLIPIQA, against 62.37 and 0.643). Increasing $\rho$ strengthens the IQA update on the detection-free images and monotonically improves perceptual quality, reaching 71.76 MUSIQ and 0.759 CLIPIQA at $\rho{=}1$, while OCR-F1 stays relatively consistent. Full-reference
fidelity declines, however: PSNR falls from 25.15 to 24.83 and LPIPS rises from 0.284 to 0.291. Our default, $\rho{=}0.3$, sits roughly halfway along this spectrum, recovering 4.44 of the 8.96 MUSIQ points and 0.047 of the 0.109 CLIPIQA points available between $\rho{=}0$ and $\rho{=}1$, while attaining the best OCR-F1 in the table.

Inference cost is reported in full in \cref{tab:efficiency}. TeReDiff and SUPIR are profiled on one NVIDIA H100 (80\,GB) over 100 Real-Text images, whereas PiSA-SR is profiled on one
NVIDIA RTX 6000 Ada (48\,GB) over all 847 Real-Text images. For TeReDiff and SUPIR, adding \our-lite (\ie, test-time adaptation in latent space with the MSE objective only) gives
a runtime overhead of $16$--$35\%$, which stems primarily from encoding the reference image, running the hallucinated text detector head at all 50 sampling steps, and applying the latent guidance update at all 50 steps. For PiSA-SR the relative overhead is higher ($+96\%$), of which about half is a one-time cost of reference-related processing ($0.028$\,s for
the SwinIR forward pass and $0.027$\,s to encode $I_{\text{ref}}$) and the other half is the single detection pass its one-step sampler allows ($0.028$\,s for the hallucinated text
detector and $0.029$\,s to recompute the mask on $z_0$). Overall, \our-lite introduces a small runtime overhead and a minimal memory overhead over the base model (at most $0.4$\,GB).

\our introduces greater runtime and memory overhead, because the NR-IQA objectives require running the VAE decoder to obtain the intermediate SR image. Compared with test-time adaptation in RGB space using an MSE-only objective (\emph{w/ RGB MSE only}), which pays for that same decode, \our adds at most $24\%$ runtime and $12\%$ memory, which is the cost of computing the NR-IQA objectives themselves (\eg, MUSIQ and CLIPIQA). In exchange we observe an improvement in both no-reference image quality and OCR performance on all three backbones. Comparing \our with \our-lite, overall OCR performance is similar (within $0.5$ OCR-F1 in every block), but \our improves the NR-IQA metrics, including those not part of the objective (QAlign $3.534$ \vs $3.446$ on TeReDiff, $3.212$ \vs $3.204$ on PiSA-SR and $3.463$ \vs $3.425$ on SUPIR).

Overall, which variant to prefer therefore depends on whether perceptual quality or inference cost is the binding constraint. \our-lite is the better default when recognition is the objective and compute is scarce; the full objective is what keeps no-reference quality from regressing below the base model, and that is what its additional cost is spent on.

\section{Results on Hallucinated Text Detector}
\label{sec:results_htd_supp}
{
\setlength{\tabcolsep}{0.04in}
\newcommand{\ra}[1]{\renewcommand{\arraystretch}{#1}}
\begin{table*}[t]
    \centering
    \fontsize{6}{7} %

    \begin{adjustbox}{max width=\linewidth}
    \begin{tabular}{@{}lrrrcrrrcrrr@{}}\toprule
      & \multicolumn{3}{c}{Real-Text-HTD-TeReDiff} & \phantom{abc}& \multicolumn{3}{c}{Real-Text-HTD-SeeSR} &
\phantom{abc} & \multicolumn{3}{c}{Real-Text-HTD-StableSR}\\
    \cmidrule{2-4} \cmidrule{6-8} \cmidrule{10-12}
       & Precision   & Recall   & Hmean   && Precision   & Recall   & Hmean   & & Precision   & Recall   & Hmean     \\ \midrule
    
    \rule{0pt}{9pt}\quad\quad TESTR       & 0.282  & 0.477   & 0.354  & & 0.242  & 0.344  & 0.284  && 0.235  & 0.304  & 0.265 \\
  \quad\quad TeReDiff$^\ddagger$       & 0.252  & 0.559   & 0.347   && 0.252  & 0.474  & 0.330  && 0.260  & 0.449  & 0.329 \\
   
Prediction confidence > 0.3 &&&&&&&&&&&\\ 
   \quad\quad TeReDiff-TeReDiff        & 0.495  & 0.340   & 0.403   && 0.467  & 0.292  &  0.359  && 0.551  & 0.289  & 0.379 \\
    \quad\quad AllModel-TeReDiff    & 0.532  & 0.333   & 0.410   && 0.508  & 0.273  &  0.355  && 0.600  & 0.292  & 0.393 \\
 \quad\quad TeReDiff-AllDeg      & 0.536  & 0.345   & 0.420  & & 0.524  & 0.266  &  0.353  && 0.574  & 0.268  & 0.365 \\
  \quad\quad AllModel-AllDeg*  & 0.601  & 0.326   & 0.422  & & 0.624  & 0.247  &  0.354 & & 0.653  & 0.255  & 0.366 \\
  Prediction confidence > 0.2&&&&&&&&&&&\\
  \quad\quad TeReDiff-TeReDiff        & 0.425  & 0.458   & 0.441   && 0.424  & 0.389  &  0.406  && 0.472  & 0.397  & 0.432 \\
 \quad\quad AllModel-TeReDiff    & 0.477  & 0.466   & 0.471   && 0.478   & 0.402  & 0.437  && 0.538  & 0.422  & 0.473 \\
    \quad\quad TeReDiff-AllDeg      & 0.454  & 0.424   & 0.438   && 0.426  & 0.331  &  0.372  && 0.474  & 0.353  & 0.404 \\
  \quad\quad AllModel-AllDeg*     & 0.524  & 0.414   & 0.462  & & 0.533  & 0.312  &  0.394  && 0.568  & 0.342  & 0.427 \\
    \bottomrule  
    \end{tabular}
    
    \end{adjustbox}
    \caption{
    \textbf{Prediction results on SR images}.
        Our hallucinated text detector shows improved detection precisions on hallucinated text regions of Real-Text-HTD across all training settings, and this improvement generalizes to images generated by SR models not seen during training (SeeSR and StableSR). While including multiple degradation during training improves detection precision by up to 7\%, it negatively affects recall.  $\ddagger$ denotes stage-3 pre-trained model from TeReDiff, * denotes the default model used in our test-time adaptation experiments for TeReDiff. }
    \label{tab:prediction-head-tair-sr}
    \end{table*}
}

\subsection{Detection Performance on Real-Text}
\cref{tab:prediction-head-tair-sr} shows evaluation results of our hallucinated text detectors under different training settings on three variations of Real-Text-HTD obtained from different SR models. Training settings are denoted in the format of <Model outputs included in training>-<degradations included in training>. For example, a hallucinated text detector trained on all SR model outputs and all degradations is denoted as AllModel-AllDeg. As shown in \cref{tab:prediction-head-tair-sr}, on Real-Text-HTD-TeReDiff, including more SR model outputs improves the precision of hallucinated text detection. By training on more degradations, precision is improved further, but at a cost of degraded recall. As detection precision is crucial for enhancing fidelity and maintaining perceptual quality at inference time, we use AllModel-AllDeg with a prediction confidence threshold of 0.3 as our default setup in test-time adaptation for TeReDiff + \our. The AllModel-AllDeg version of our model shows strong detection precisions across various evaluation settings.
\subsection{Detection Performance on SA-Text-Test}
{
\setlength{\tabcolsep}{0.04in}
\newcommand{\ra}[1]{\renewcommand{\arraystretch}{#1}}
\begin{table*}[t]
    \centering
    \fontsize{6}{7} %

    \begin{adjustbox}{max width=\linewidth}
    \begin{tabular}{@{}lrrrcrrrcrrr@{}}\toprule
      & \multicolumn{3}{c}{SA-Text-Test-HTD-TeReDiff} & \phantom{abc}& \multicolumn{3}{c}{SA-Text-Test-HTD-SeeSR} &
\phantom{abc} & \multicolumn{3}{c}{SA-Text-Test-HTD-StableSR}\\
    \cmidrule{2-4} \cmidrule{6-8} \cmidrule{10-12}
       & Precision   & Recall   & Hmean   && Precision   & Recall   & Hmean   & & Precision   & Recall   & Hmean     \\ \midrule
   Prediction confidence > 0.3 &&&&&&&&&&&\\
    \quad\quad TeReDiff-TeReDiff        & 0.688 & 0.468 & 0.557  && 0.643 & 0.449 & 0.529  && 0.628 & 0.424 & 0.506 \\
   \quad\quad AllModel-TeReDiff    & 0.743 & 0.443 & 0.555  && 0.709 & 0.421 & 0.528  && 0.710 & 0.382 & 0.497 \\
   \quad\quad TeReDiff-AllDeg      & 0.700 & 0.429 & 0.532  & & 0.626 & 0.361 & 0.458  && 0.655 & 0.344 & 0.451 \\
   \quad\quad AllModel-AllDeg*  & 0.759 & 0.427 & 0.546  & & 0.729 & 0.331 & 0.455 & & 0.743 & 0.335 & 0.462 \\
  Prediction confidence > 0.2&&&&&&&&&&&\\
  \quad\quad TeReDiff-TeReDiff        & 0.648 & 0.569 & 0.606   && 0.596 & 0.532 & 0.562 && 0.584 & 0.495 & 0.536 \\
  \quad\quad AllModel-TeReDiff    & 0.680 & 0.540 & 0.602   && 0.665 & 0.512 & 0.579  && 0.656 & 0.471 & 0.548 \\
   \quad\quad TeReDiff-AllDeg      & 0.647 & 0.488 & 0.556   && 0.570 & 0.398 & 0.469 && 0.606 & 0.388 & 0.473 \\
  \quad\quad AllModel-AllDeg*     & 0.698 & 0.471 & 0.562 & & 0.679 & 0.386 & 0.493  && 0.695 & 0.379 & 0.490 \\
    \bottomrule  
    \end{tabular}
    
    \end{adjustbox}
    \caption{
    \textbf{Prediction results on SA-Text-Test.}
        We include evaluation of multiple training settings on SA-Text-Test-HTD test set created following procedures described in \cref{sec:test_set_creation}. Training settings (rows) are denoted in the format of <Model outputs included in training>-<degradations included in training>. We observe a similar trend as in \cref{tab:prediction-head-tair-sr}, with the AllModel-AllDeg setting achieving the highest detection precision. * denotes the default model used in our test-time adaptation experiments for TeReDiff.}
    
    \label{tab:prediction-head-sa-test}
    \end{table*}
}

We include detection performance on the SA-Text-Test dataset with Level 2 degradation in \cref{tab:prediction-head-sa-test}. We observe a similar trend as \cref{tab:prediction-head-tair-sr} in precision: it benefits from including more SR model outputs and more degradations in training. In our task, precision is particularly important, as false hallucination predictions could lead to legible text being incorrectly suppressed. Such errors will directly harm readability and recognition recall. 

\section{Additional Qualitative Examples}
\label{sec:quali_supp}

\begin{figure*}[tb]
  \centering
  \begin{minipage}{0.9\linewidth}
    \includegraphics[width=\linewidth]{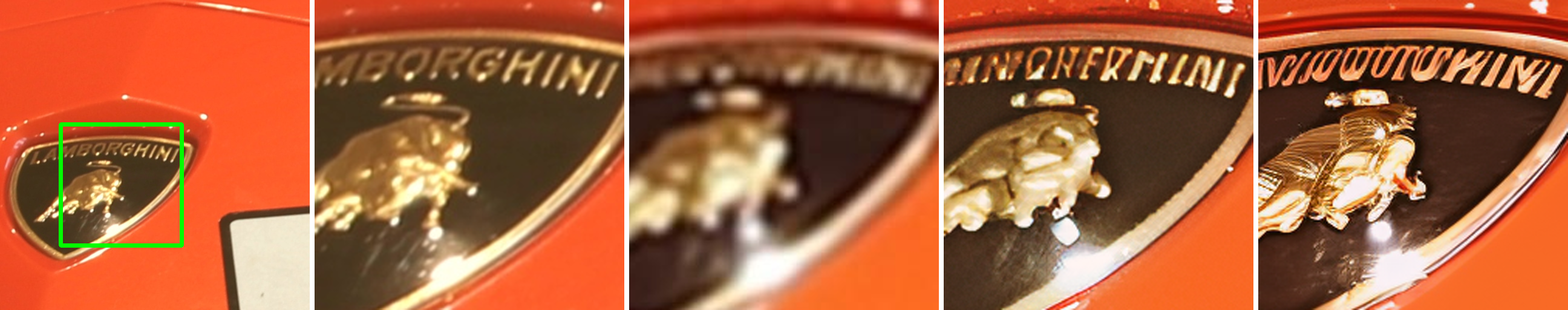}
    \vfill \vspace*{-0.12cm}
    \text{ ~~~~~~~~~~~~~~~GT~~~~~~~~~~~~~~~~~~~~~~~~~~~~~~~GT~~~~~~~~~~~~~~~~~~~~~~~~~~~~Bicubic~~~~~~~~~~~~~~~~~~~~~~~StableSR~~~~~~~~~~~~~~~~~~~~~~~SeeSR}
    \vfill \vspace*{0.12cm}
  \end{minipage}
  \vfill
  \begin{minipage}{0.9\linewidth}
    \includegraphics[width=\linewidth]{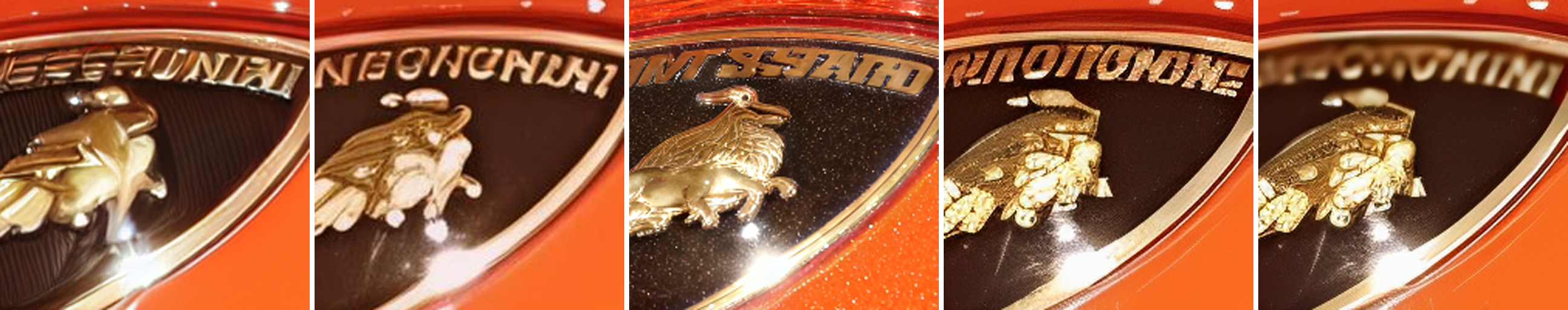}
    \vfill \vspace*{-0.12cm}
\text{~~~~~~~~~~~~~~~PASD~~~~~~~~~~~~~~~~~~~~~~~PiSA-SR~~~~~~~~~~~~~~~~~~~~~~SUPIR~~~~~~~~~~~~~~~~~~~~~~~~~TeReDiff~~~~~~~~~~~~~~~TeReDiff + Ours}
    \vfill \vspace*{0.12cm}
  \end{minipage}

  \vfill
  \begin{minipage}{0.9\linewidth}
    \includegraphics[width=\linewidth]{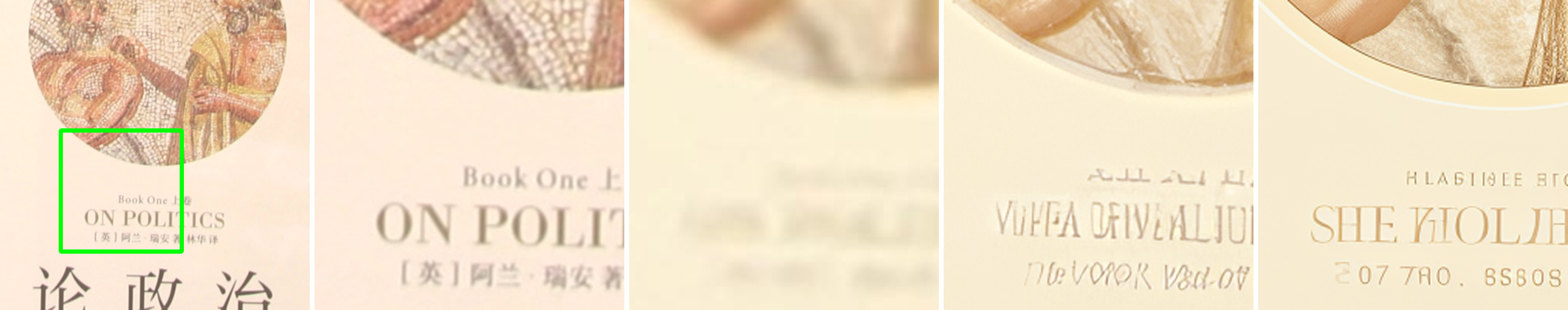}
    \vfill \vspace*{-0.12cm}
    \text{ ~~~~~~~~~~~~~~~GT~~~~~~~~~~~~~~~~~~~~~~~~~~~~~~~GT~~~~~~~~~~~~~~~~~~~~~~~~~~~~Bicubic~~~~~~~~~~~~~~~~~~~~~~~StableSR~~~~~~~~~~~~~~~~~~~~~~~SeeSR}
    \vfill \vspace*{0.12cm}
  \end{minipage}
  \vfill
  \begin{minipage}{0.9\linewidth}
    \includegraphics[width=\linewidth]{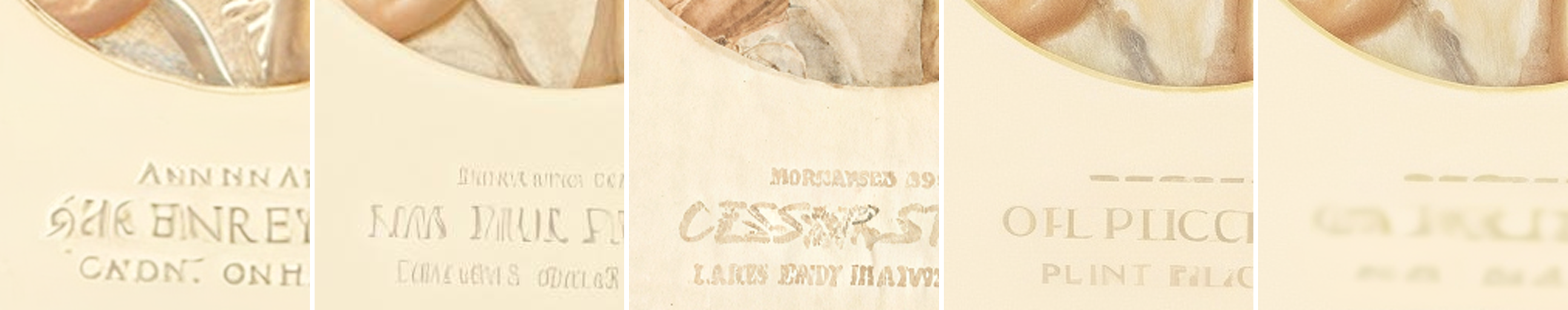}
    \vfill \vspace*{-0.12cm}
    \text{~~~~~~~~~~~~~~~PASD~~~~~~~~~~~~~~~~~~~~~~~PiSA-SR~~~~~~~~~~~~~~~~~~~~~~SUPIR~~~~~~~~~~~~~~~~~~~~~~~~~TeReDiff~~~~~~~~~~~~~~~TeReDiff + Ours}
    \vfill \vspace*{0.12cm}
  \end{minipage}
  
  \caption{\textbf{Qualitative results on the Real-Text dataset}. We show two examples. For each example, the first image is the GT image with a bounding box defining a cropped region containing text, and the rest are the crops in GT and outputs from the various SR methods. When the input image lacks sufficient visual cues, diffusion-based SR models tend to hallucinate text in the restored regions. Our approach (TeReDiff + Ours) effectively suppresses these hallucinations while preserving plausible structures. %
  }
  \label{fig:tgsr_qual_real_text_supp3}
\end{figure*}

\begin{figure*}[tb]
  \centering
  \begin{minipage}{0.9\linewidth}
    \includegraphics[width=\linewidth]{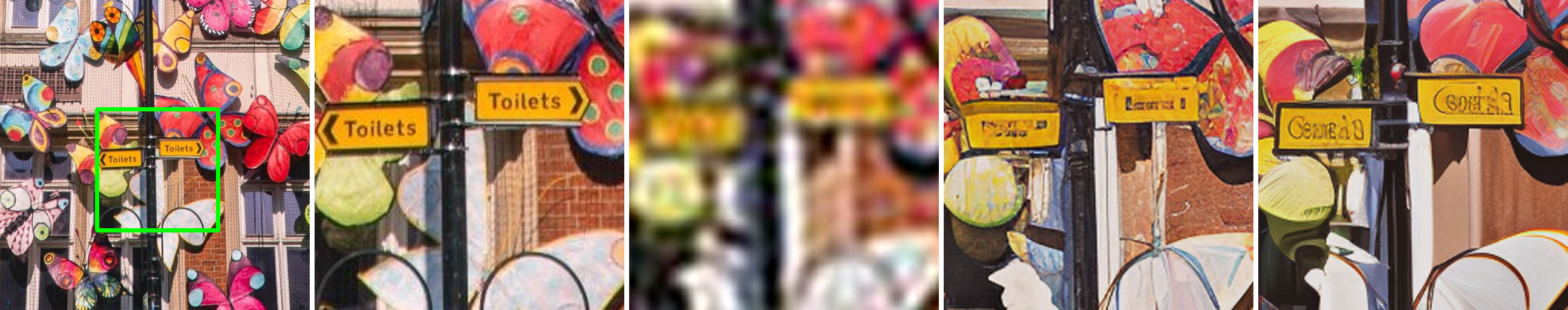}
    \vfill \vspace*{-0.12cm}
    \text{ ~~~~~~~~~~~~~~~GT~~~~~~~~~~~~~~~~~~~~~~~~~~~~~~~GT~~~~~~~~~~~~~~~~~~~~~~~~~~~~Bicubic~~~~~~~~~~~~~~~~~~~~~~~StableSR~~~~~~~~~~~~~~~~~~~~~~~SeeSR}
    \vfill \vspace*{0.12cm}
  \end{minipage}
  \vfill
  \begin{minipage}{0.9\linewidth}
    \includegraphics[width=\linewidth]{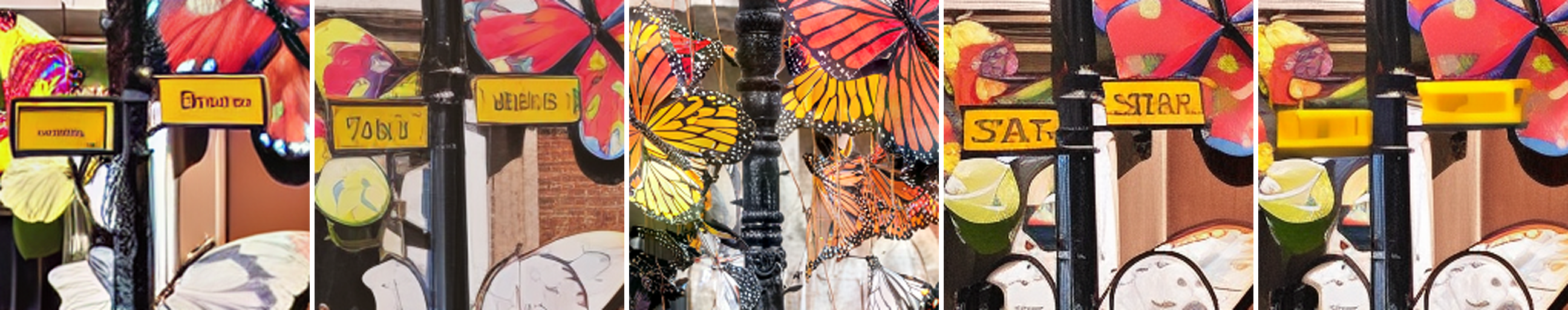}
    \vfill \vspace*{-0.12cm}
    \text{~~~~~~~~~~~~~~~PASD~~~~~~~~~~~~~~~~~~~~~~~PiSA-SR~~~~~~~~~~~~~~~~~~~~~~SUPIR~~~~~~~~~~~~~~~~~~~~~~~~~TeReDiff~~~~~~~~~~~~~~~TeReDiff + Ours}
    \vfill \vspace*{0.12cm}
  \end{minipage}
  \vfill
  \begin{minipage}{0.9\linewidth}
    \includegraphics[width=\linewidth]{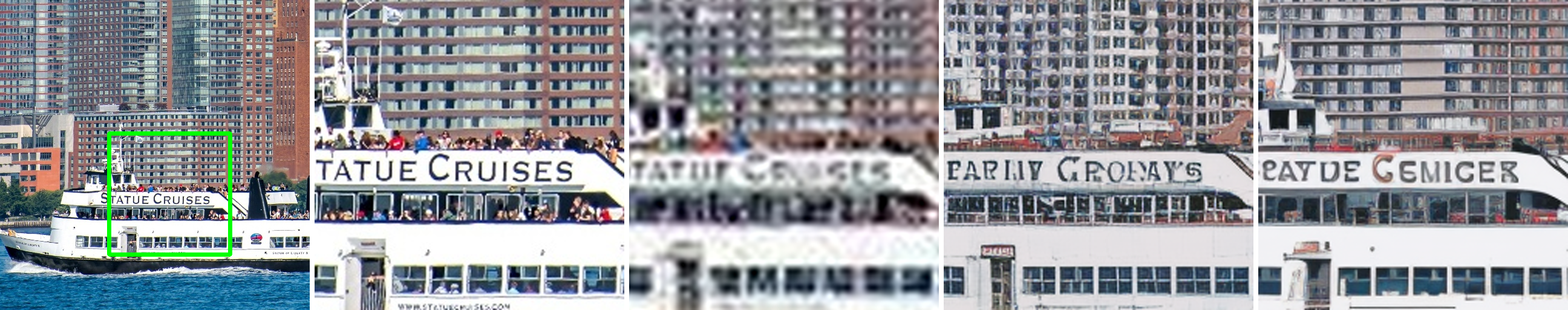}
    \vfill \vspace*{-0.12cm}
    \text{ ~~~~~~~~~~~~~~~GT~~~~~~~~~~~~~~~~~~~~~~~~~~~~~~~GT~~~~~~~~~~~~~~~~~~~~~~~~~~~~Bicubic~~~~~~~~~~~~~~~~~~~~~~~StableSR~~~~~~~~~~~~~~~~~~~~~~~SeeSR}
    \vfill \vspace*{0.12cm}
  \end{minipage}

  \vfill
  \begin{minipage}{0.9\linewidth}
    \includegraphics[width=\linewidth]{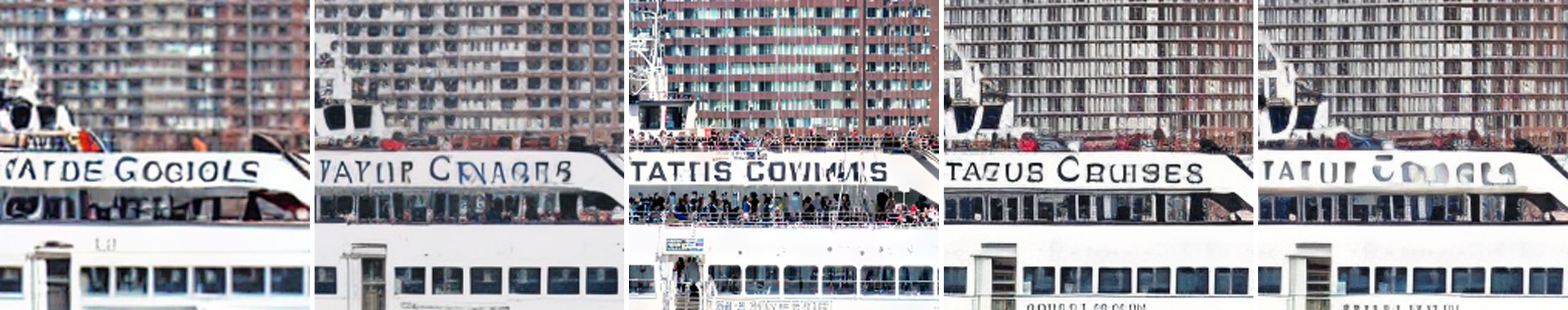}
    \vfill \vspace*{-0.12cm}
    \text{~~~~~~~~~~~~~~~PASD~~~~~~~~~~~~~~~~~~~~~~~PiSA-SR~~~~~~~~~~~~~~~~~~~~~~SUPIR~~~~~~~~~~~~~~~~~~~~~~~~~TeReDiff~~~~~~~~~~~~~~~TeReDiff + Ours}
    \vfill \vspace*{0.12cm}
  \end{minipage}
  
  \vfill
  \begin{minipage}{0.9\linewidth}
    \includegraphics[width=\linewidth]{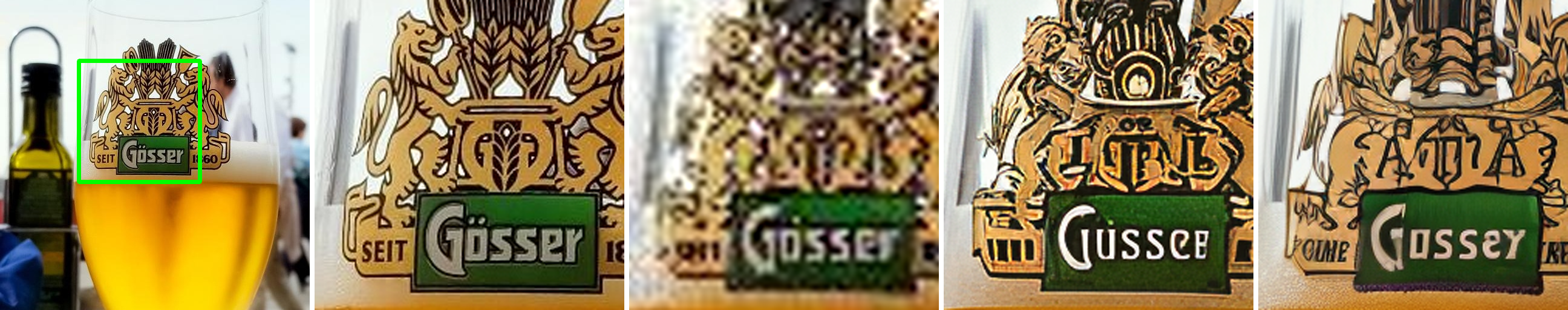}
    \vfill \vspace*{-0.12cm}
    \text{ ~~~~~~~~~~~~~~~GT~~~~~~~~~~~~~~~~~~~~~~~~~~~~~~~GT~~~~~~~~~~~~~~~~~~~~~~~~~~~~Bicubic~~~~~~~~~~~~~~~~~~~~~~~StableSR~~~~~~~~~~~~~~~~~~~~~~~SeeSR}
    \vfill \vspace*{0.12cm}
  \end{minipage}
  \vfill
  \begin{minipage}{0.9\linewidth}
    \includegraphics[width=\linewidth]{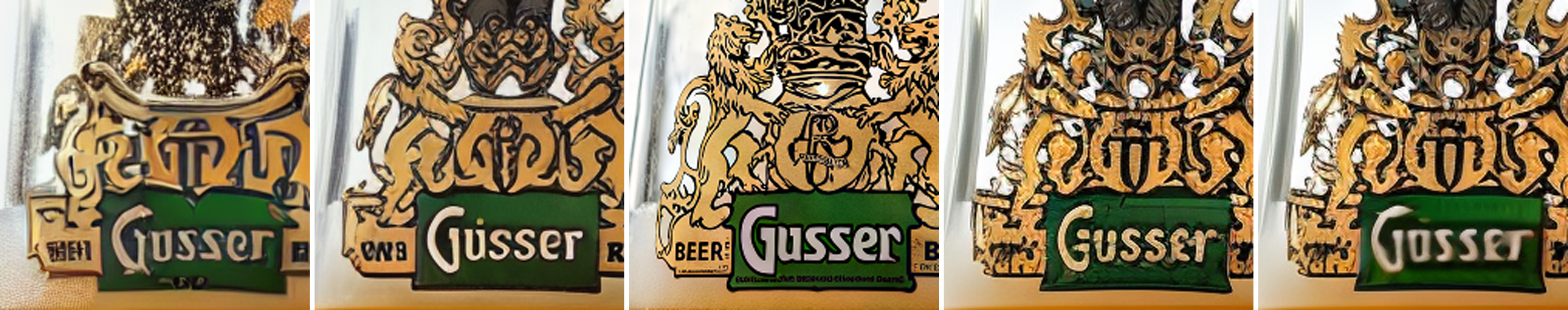}
    \vfill \vspace*{-0.12cm}
   \text{~~~~~~~~~~~~~~~PASD~~~~~~~~~~~~~~~~~~~~~~~PiSA-SR~~~~~~~~~~~~~~~~~~~~~~SUPIR~~~~~~~~~~~~~~~~~~~~~~~~~TeReDiff~~~~~~~~~~~~~~~TeReDiff + Ours}
    \vfill \vspace*{0.12cm}
  \end{minipage}
  
  \caption{\textbf{Qualitative results on the SA-Text-Test dataset}. We show three examples. For each example, the first image is the GT image with a bounding box defining a cropped region containing text, and the rest are the crops in GT and outputs from the various SR methods. When the input image lacks sufficient visual cues, diffusion-based SR models tend to hallucinate text in the restored regions. Our approach (TeReDiff + Ours) effectively suppresses these hallucinations while preserving plausible structures. %
  }
  \label{fig:tgsr_qual_sa_text_supp3}
\end{figure*}

\begin{figure*}[tb]
  \centering
  \begin{minipage}{0.9\linewidth}
    \text{ ~~~~~~~~~~~~~~~GT~~~~~~~~~~~~~~~~~~~~~~~~~~~~~~~GT~~~~~~~~~~~~~~~~~~~~~~~~~~~~TeReDiff~~~~~~~~~~TeReDiff + Ours(Lite)~~~~~TeReDiff + Ours}
    \vfill
    \includegraphics[width=\linewidth]{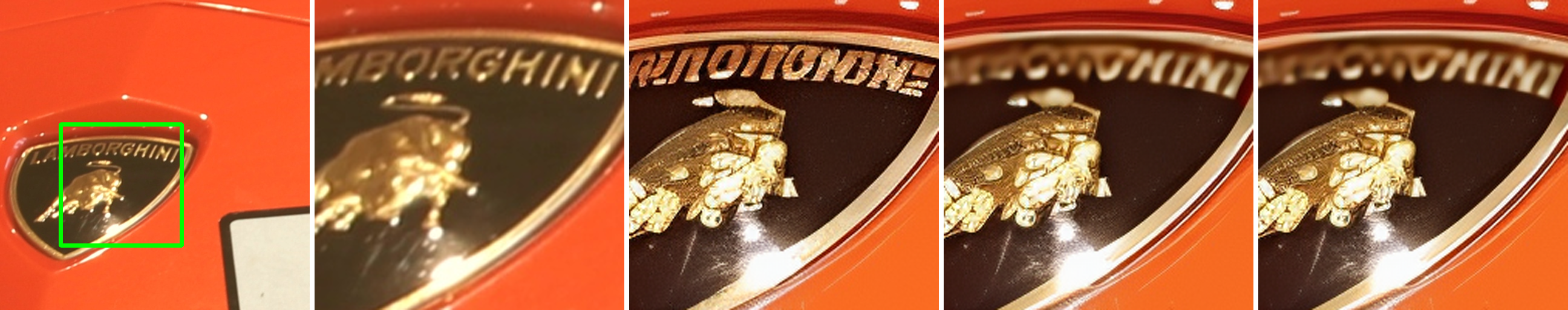}
  \end{minipage}
  \vfill
    \begin{minipage}{0.9\linewidth}
    \includegraphics[width=\linewidth]{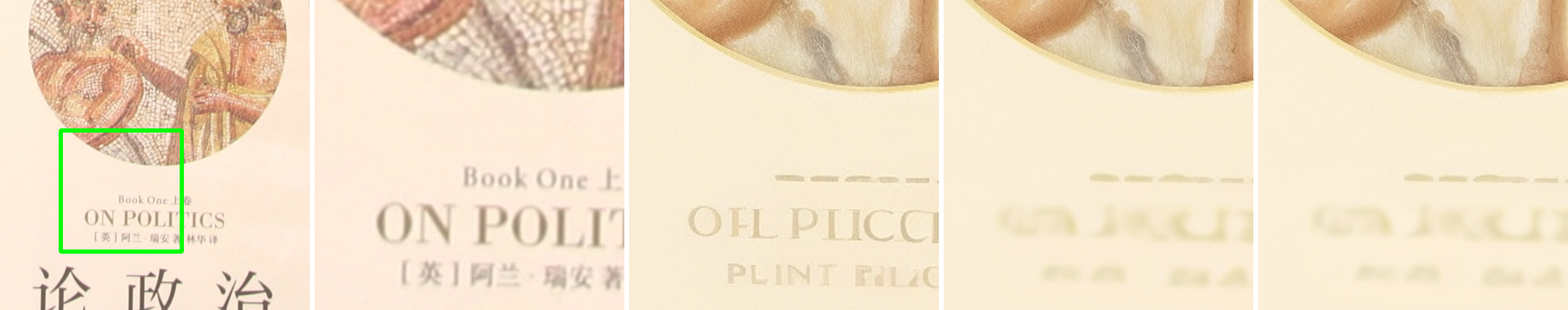}
  \end{minipage}
    \vfill\vspace{5mm}
  \begin{minipage}{0.9\linewidth}
    \includegraphics[width=\linewidth]{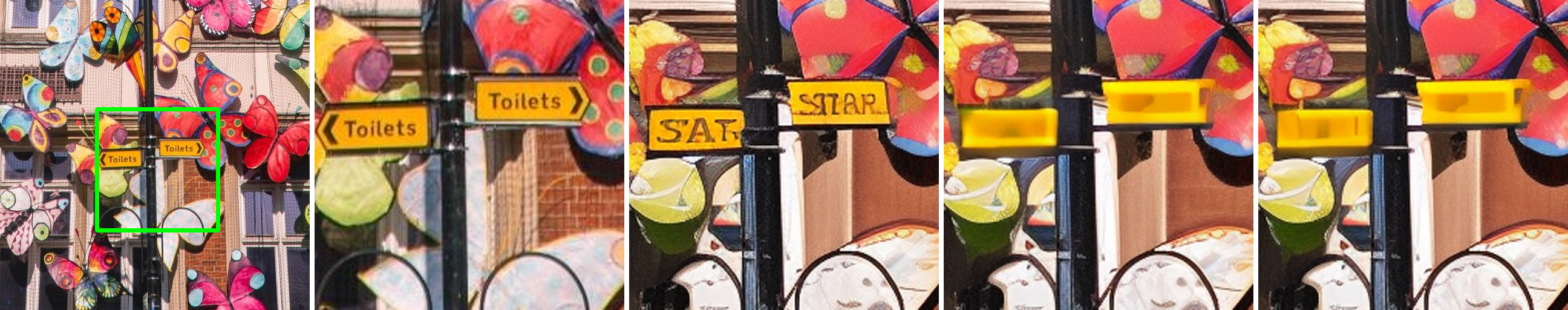}
  \end{minipage}
  \vfill
    \begin{minipage}{0.9\linewidth}
    \includegraphics[width=\linewidth]{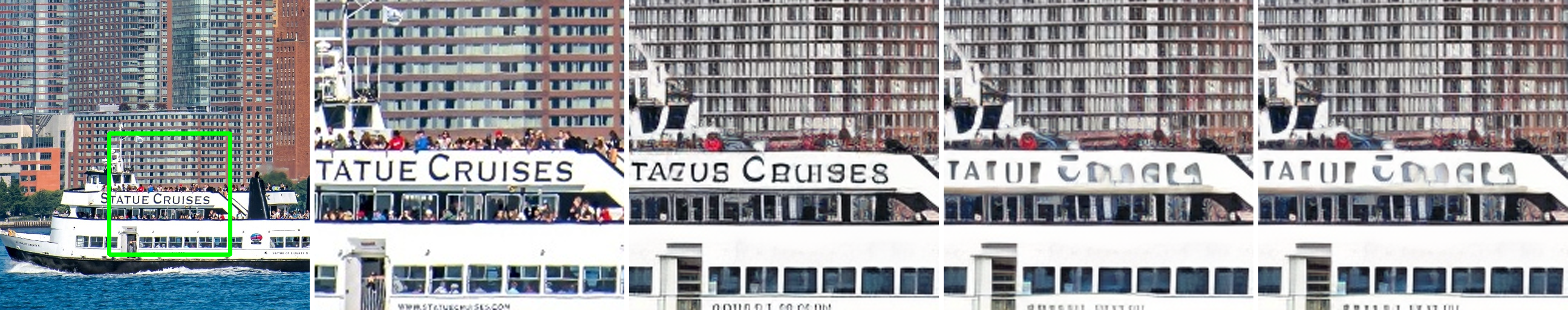}
  \end{minipage}
    \vfill
    \begin{minipage}{0.9\linewidth}
    \includegraphics[width=\linewidth]{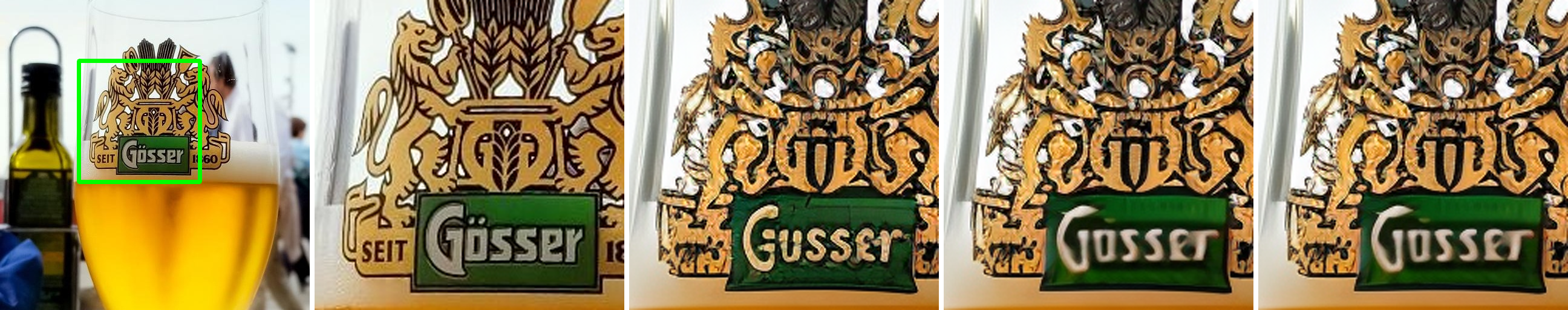}
  \end{minipage}
  \caption{\textbf{Qualitative results on the Real-Text (top two rows) and SA-Text-Test (bottom three rows) datasets}. For each example, the first image is the GT image with a bounding box defining a cropped region containing text, and the rest are the crops in GT and outputs from the TeReDiff and our two models. Compared to TeReDiff, our approaches (TeReDiff + Ours(Lite))  and (TeReDiff + Ours) effectively suppress hallucinations while preserving plausible structures. Our lite model (TeReDiff + Ours(Lite)) exhibits competitive performance. 
  }
  \label{fig:tgsr_qual_lite}
\end{figure*}

We provide additional qualitative examples on the \textbf{Real-Text} dataset (\cref{fig:tgsr_qual_real_text_supp3}) and the \textbf{SA-Text-Test} dataset (\cref{fig:tgsr_qual_sa_text_supp3}). As shown in these examples, existing methods often suffer from severe text hallucinations and generate distorted text in many cases. In contrast, our approach effectively suppresses such hallucinations while preserving the perceptual quality of legible text and background regions. Furthermore, we compare \our with \our-lite in \cref{fig:tgsr_qual_lite}. Building upon the efficiency gains previously mentioned (see~\cref{tab:efficiency}), we now demonstrate that \our-lite maintains visual quality comparable to that of \our.

\end{document}